\documentclass{article}

\usepackage[preprint]{neurips_2026}

\usepackage{enumitem}
\usepackage{times}
\usepackage{soul}
\usepackage{url}
\usepackage[hidelinks]{hyperref}
\usepackage[utf8]{inputenc}
\usepackage[small]{caption}
\usepackage{graphicx}
\usepackage{amsmath}
\usepackage{amsthm}
\usepackage{amssymb}
\usepackage{booktabs}
\usepackage{algorithm}
\usepackage{algorithmic}
\usepackage{listings}
\lstdefinelanguage{json}{
  basicstyle=\ttfamily,
  showstringspaces=false,
  breaklines=true,
  breakatwhitespace=false,
  columns=fullflexible,
  keepspaces=true,
  morestring=[b]",
  stringstyle=\ttfamily,
  literate=
   *{0}{{0}}1 {1}{{1}}1 {2}{{2}}1 {3}{{3}}1 {4}{{4}}1
    {5}{{5}}1 {6}{{6}}1 {7}{{7}}1 {8}{{8}}1 {9}{{9}}1
    {:}{{:}}1 {,}{{,}}1 {\{}{{\{}}1 {\}}{{\}}}1 {[}{{[}}1 {]}{{]}}1
}
\usepackage{xcolor}
\usepackage{wrapfig}
\usepackage{multirow}

\newcommand{\method}[0]{\text{LexGuard}}

\title{Which Changes Matter? Towards Trustworthy Legal AI via Relevance-Sensitive Evaluation and Solver-Grounded Reasoning}

\author{
Chen Linze$^{1}$, Cai Yufan$^{1}$, Hou Zhe$^{2}$ and Dong Jin Song$^{1}$\\
$^{1}$National University of Singapore\\
$^{2}$Griffith University\\
\begin{tabular}{c}
\texttt{e1344652@u.nus.edu, cai\_yufan@u.nus.edu,}\\
\texttt{z.hou@griffith.edu.au, dcsdjs@nus.edu.sg}
\end{tabular}
}

\begin{document}
\maketitle

\begin{abstract}
Legal reasoning requires distinguishing changes that matter from those that do not.
Legal AI should remain stable under legally irrelevant perturbations, but should change when perturbations alter legally material points.
We formulate this requirement as a legal-relevance-sensitive evaluation problem: LLMs should only be sensitive to the legally relevant change.
We introduce a unified evaluation suite covering \emph{should-change} and \emph{should-not-change} evaluation across judicial fairness, robustness, and statute-confusion scenarios. 
Our evaluation shows that existing legal LLMs are systematically sensitive to legally irrelevant variations and often fail to distinguish related legal elements and statutory rules. 
To mitigate these failures, we present \method, an adversarial multi-agent framework grounded in formal reasoning.
\method{} formalizes statutes into executable constraints, uses adversarial agents to extract competing fact--statute arguments, and invokes SMT solvers to verify legal satisfaction and logical consistency. 
Experiments show that \method{} improves legal reasoning reliability by reducing vulnerability to manipulative framing, improving disambiguation among similar statutes, limiting the influence of legally irrelevant attributes, and increasing consistency under benign reformulations.
We show that legal trustworthiness requires not only accuracy, but calibrated sensitivity to legally material changes.
\end{abstract}

\section{Introduction}

Large language models (LLMs) are increasingly used for legal tasks such as legal question answering, opinion summarization, pleading drafting, and bar-exam-style reasoning~\cite{katz2023gpt,hendrycks2021measuring}. 
However, legal reasoning is a high-stakes setting where accuracy alone is insufficient. 
In rule-of-law systems, judicial outcomes are expected to satisfy \emph{formal rationality}: conclusions should be justified by explicit, general, and logically coherent rules~\cite{linna2025judicial,sadowski2025verifiable}, and traceable to governing statutes, interpretations, and precedents~\cite{kesari2024legal}. 
Current LLM-based legal systems, including domain-adapted models such as \textsc{ChatLaw}~\cite{cui2023chatlaw}, \textsc{LawLLM}~\cite{zheng2024lawllm}, and \textsc{Lexi}~\cite{lehmberg2025lexi}, still remain vulnerable to hallucinated authorities~\cite{chen2023hallucination}, confusion among similar statutes~\cite{savelka2023when}, and sensitivity to irrelevant attributes.

A central requirement of legal reasoning is deciding \emph{which changes matter}. 
Changes to statutory elements, mental state, harm severity, or applicable exceptions may properly alter the outcome. 
In contrast, decisions should remain invariant to legally irrelevant perturbations, such as demographic attributes, procedural background, stylistic reformulation, irrelevant expert opinions, adversarial framing, or misleading references to inapplicable statutes~\cite{yiranllms}. 
This ability to distinguish legally relevant from legally irrelevant changes is fundamental to rule-of-law values: similar cases should be treated alike, different cases should be distinguished for legally grounded reasons, and legal conclusions should be traceable to explicit and coherent legal rules.

Existing evidence suggests that current LLMs often fail precisely at this distinction~\cite{hu2025j}. 
Recent judicial fairness evaluation shows that LLMs acting as judges exhibit pervasive inconsistency, bias, and imbalanced inaccuracy across demographic, substantive, and procedural factors.
LLMs can also be misled by perturbations to the major premise, minor premise, and conclusion-generation stages of legal syllogistic reasoning, even when the legally operative facts remain unchanged~\cite{yiranllms}.

However, existing evaluations usually study these phenomena in isolation. 
We formulate fair and robust legal reasoning as a \emph{legal-relevance-sensitive evaluation} problem. 
Instead of evaluating only whether LLMs remain unchanged, we ask whether LLMs' reasoning changes under \emph{should-change} perturbations and remains stable under \emph{should-not-change} perturbations. 
We also unify judicial fairness, robustness, adversarial framing, and statute-confusion evaluation.
Based on this view, we introduce a new evaluation suite covering four perturbation families: 
(1) fairness perturbations over legally relevant or irrelevant attributes; 
(2) robustness perturbations that preserve or change legal meaning; 
(3) adversarial framing perturbations that attempt to manipulate the conclusion; and 
(4) statute-confusion perturbations involving similar legal rules and elements. 

Our evaluations show that current LLMs are often unstable under the defined perturbations and attacks.
To mitigate these failures, we propose \method, a solver-grounded adversarial multi-agent framework for legal reasoning. 
\method{} first autoformalizes statutory provisions into executable constraints capturing legal elements. 
It then uses prosecutor- and defense-aligned adversarial agents to extract structured fact--statute argument tuples independently from case narratives. 
Finally, an SMT solver checks whether the competing arguments satisfy the formalized legal constraints and whether their conclusions are logically consistent. 
This design turns relevance-sensitive legal reasoning into a auditable process: legally material changes should modify the satisfiability of the corresponding legal constraints, whereas irrelevant changes should not.

We evaluate \method{} on public legal datasets~\cite{li2023lecardv2,xue2024leec} and our relevance-sensitive evaluation suite. 
Experiments demonstrate improvements along four dimensions: 
standard legal reasoning performance, including statute selection, verdict prediction, and sentencing quality; 
fairness under legally relevant and irrelevant perturbations; 
robustness under benign reformulations and adversarial framing attacks; and 
resistance to statute-confusion errors. 
Beyond performance gains, \method{} produces solver-checked symbolic justifications, enabling legal conclusions to be audited against explicit statutory constraints.

Our contributions are summarized as follows:
\begin{itemize}
    \item We formulate trustworthy legal reasoning as a \emph{legal-relevance-sensitive evaluation} problem and instantiate it with a unified suite of \emph{should-change} and \emph{should-not-change} perturbations spanning fairness, robustness, adversarial attacks, and statute-confusion scenarios.

    \item We propose \method, a solver-grounded adversarial multi-agent framework that formalizes statutes into executable constraints, extracts competing legal arguments through adversarial agents, and check the judgment with an SMT solver.

    \item We empirically show that existing LLM legal reasoners are sensitive to legal element changes and vulnerable to statute-confusion and attacks, while \method{} improves verdict accuracy, statute selection, sentencing quality, fairness, robustness, and trustworthiness.
\end{itemize} 
\section{Legal-Relevance-Sensitive Evaluation}
\label{sec:evaluation}

\begin{table*}[t]
\centering
\footnotesize
\setlength{\tabcolsep}{5pt}
\renewcommand{\arraystretch}{1.18}
\caption{
Overview of the proposed legal-relevance-sensitive evaluation framework.
Each evaluation axis corresponds to a class of perturbations targeting a specific component of legal reasoning.
}
\label{tab:legal_relevance_eval}
\begin{tabular}{p{1.2cm} p{1.2cm} p{6.4cm} p{3.4cm}}
\toprule
\textbf{Evaluation axis}
& \textbf{Reasoning target}
& \textbf{Perturbation examples}
& \textbf{Metrics} \\
\midrule

\multicolumn{4}{l}{\textit{Label-preserving perturbations: legally irrelevant changes should not alter the decision}} \\
\midrule

Judicial fairness
& Extra-legal factors
& Demographic and procedural counterfactuals, including defendant gender, ethnicity, wealth, education, household registration, victim attributes, defender attributes, court level, trial publicity, and procedural background.
& Invariance; Bias Magnitude \\ \midrule

Benign robustness
& Surface form of case facts
& Meaning-preserving reformulations, including paraphrase, synonym substitution, reordered descriptions, stylistic rewriting, expression-level noise, and irrelevant background narration.
& Overall Score; Invariance; Bias Magnitude \\ \midrule

Major-premise robustness
& Governing legal rule
& Similar but inapplicable statutes, misleading legal references, fabricated legal authority, wrong charge names, and irrelevant retrieved provisions.
& ASR; CRR; Invariance; Attack-aware F1 \\ \midrule

Minor-premise robustness
& Legally relevant fact extraction
& Non-material factual edits, legally equivalent descriptions of elements, irrelevant factual additions, narration changes, and confusing but non-dispositive factual details.
& Overall Score; Invariance; Bias Magnitude \\ \midrule

Conclusion-level robustness
& Final legal conclusion
& Irrelevant expert opinions, prior unrelated behavior, emotionally loaded framing, role hijacking, verdict-forcing instructions, and format-mimicking prompt injection.
& ASR; CRR; Invariance; Attack-aware F1 \\  
 
\midrule
\multicolumn{4}{l}{\textit{Label-changing perturbations: legally material changes should alter the decision}} \\
\midrule

Statutory-element sensitivity
& Constitutive legal elements
& Changes to conduct, object, subject identity, role in offense, amount threshold, harm severity, causation, or other offense-defining facts.
& Overall Score; Change Alignment; Statute Correctness; Bias Magnitude \\ \midrule

Mental-state sensitivity
& Subjective culpability
& Intentional vs negligent conduct, knowing vs unknowing participation, purpose of illegal possession, awareness of illegality, or intent to cause harm. 
& Overall Score; Change Alignment; Statute Correctness; Bias Magnitude \\ \midrule

Exception and condition sensitivity
& Legal applicability conditions
& Self-defense, attempt, accomplice status, surrender, recidivism, statutory mitigation, aggravating circumstances, or other exceptions and sentencing conditions.
& Overall Score; Change Alignment; Statute Correctness; Bias Magnitude \\ \midrule

Statute-confusion sensitivity
& Boundary between related statutes
& Confusing charges or provisions with overlapping surface facts, such as theft vs fraud, robbery vs extortion, intentional injury vs intentional homicide, concealment vs money laundering, or other similar-statute clusters.
& Positive Exactness; Macro Exactness; Gold Omission; Wrong Similar Selection \\

\bottomrule
\end{tabular}
\end{table*}

A trustworthy legal model should remain stable under changes that are irrelevant, yet update its decision when a perturbation changes statutory elements, exceptions, or legal consequences. 
We propose a \emph{legal-relevance-sensitive evaluation} framework that unifies counterfactual fairness evaluation and reasoning-chain robustness evaluation.
Table~\ref{tab:legal_relevance_eval} summarizes the evaluation design.
The upper block contains label-preserving perturbations, where the model should remain invariant.
The lower block contains label-changing perturbations, where the model should update its prediction.

\subsection{Evaluation Principle}

Formally, let $x$ denote an original case, $f$ a legal reasoning model, and $\tau$ a perturbation operator that produces a modified case $\tau(x)$. 
The model produces:
$
y = f(x), y' = f(\tau(x)).
$
The central question is not simply whether $y$ and $y'$ are identical, but whether the change from $y$ to $y'$ is legally justified.
We classify perturbations into two categories. 
A \emph{label-preserving perturbation} changes only legally irrelevant information while preserving the material facts and applicable law. 
For these perturbations, a legally grounded model should satisfy:
$
f(x) = f(\tau(x)).
$
A \emph{label-changing perturbation} modifies at least one legally material condition.
For these perturbations, the model should update its prediction according to the new legal label:
$
f(\tau(x)) = y_{\tau},
$
where $y_{\tau}$ denotes the new label after the legally material modification.
For label-preserving perturbations, we measure whether the model remains invariant:
$
\mathrm{Inv}(f)
=
\mathbb{E}_{(x,\tau)}
\left[
\mathbb{I}\{f(x)=f(\tau(x))\}
\right].
$
Lower invariance indicates that the model is sensitive to legally irrelevant changes.
For label-changing perturbations, we measure change alignment: 
$
\mathrm{Align}(f)
=
\mathbb{E}_{(x,\tau)}
\left[
\mathbb{I}\{f(\tau(x))\neq f(x)\}
\right].
$
More details are in the Appendix(\ref{app:metrics}).

\subsection{Perturbation Taxonomy}

As summarized in Table~\ref{tab:legal_relevance_eval}, our taxonomy covers extra-legal factors, surface factual expression, legal-rule selection, fact extraction, conclusion generation, statutory elements, legal applicability conditions, and boundaries between related statutes.

\paragraph{Label-preserving perturbations.}
These perturbations preserve the legally material facts and applicable law. 
\emph{Judicial fairness} targets extra-legal factors such as defendant demographics, victim attributes, defender attributes, court level, trial publicity, and procedural background. 
\emph{Benign robustness} targets the surface form of facts through paraphrases, synonym substitutions, reordered descriptions, stylistic rewriting, and irrelevant narration. 
\emph{Major-premise robustness} targets the governing legal rule by injecting similar but inapplicable statutes, misleading references, fabricated authorities, wrong charge names, or irrelevant retrieved provisions. 
\emph{Minor-premise robustness} targets fact extraction through non-dispositive factual edits, legally equivalent element descriptions, irrelevant factual additions, and confusing but immaterial details. 
\emph{Conclusion-level robustness} targets final decision generation by adding irrelevant expert opinions, prior unrelated behavior, emotional framing, role hijacking, verdict-forcing instructions, or format-mimicking prompt injections. 

\paragraph{Label-changing perturbations.}
These perturbations alter legally material conditions and therefore require the model to update its prediction. 
\emph{Statutory-element sensitivity} changes constitutive elements such as conduct, object, subject identity, role in offense, amount threshold, harm severity, or causation. 
\emph{Mental-state sensitivity} modifies subjective culpability, such as intent, negligence, knowledge, purpose, or awareness of illegality. 
\emph{Exception and condition sensitivity} changes legal applicability conditions, including self-defense, attempt, accomplice status, surrender, recidivism, mitigation, or aggravation. 
\emph{Statute-confusion sensitivity} tests boundaries between related statutes with overlapping surface facts but different applicability conditions. 

\subsection{Evaluation Protocol}

For each original case, we construct paired perturbation cases. 
Each perturbation is annotated with two types of metadata: 
(i) whether it is label-preserving or label-changing, and 
(ii) which reasoning component it targets. 
For label-preserving perturbations, the gold label remains unchanged. 
For label-changing perturbations,  the gold label should reflect the legally modified outcome.
We then evaluate each model on both original and perturbed cases. 
The outputs are compared at multiple levels, including final verdict, applicable statute set, general-provision selection, specific-provision selection, and sentencing result when available. 
This protocol enables us to diagnose four distinct failure modes: unfairness under extra-legal counterfactuals, instability under benign reformulations, susceptibility to adversarial legal framing, and inability to distinguish legally material from immaterial statutory changes.
\section{Solver-grounded Reasoning}
\label{sec:approach}

\begin{figure*}[t]
    \centering
    \includegraphics[width=1.0\linewidth]{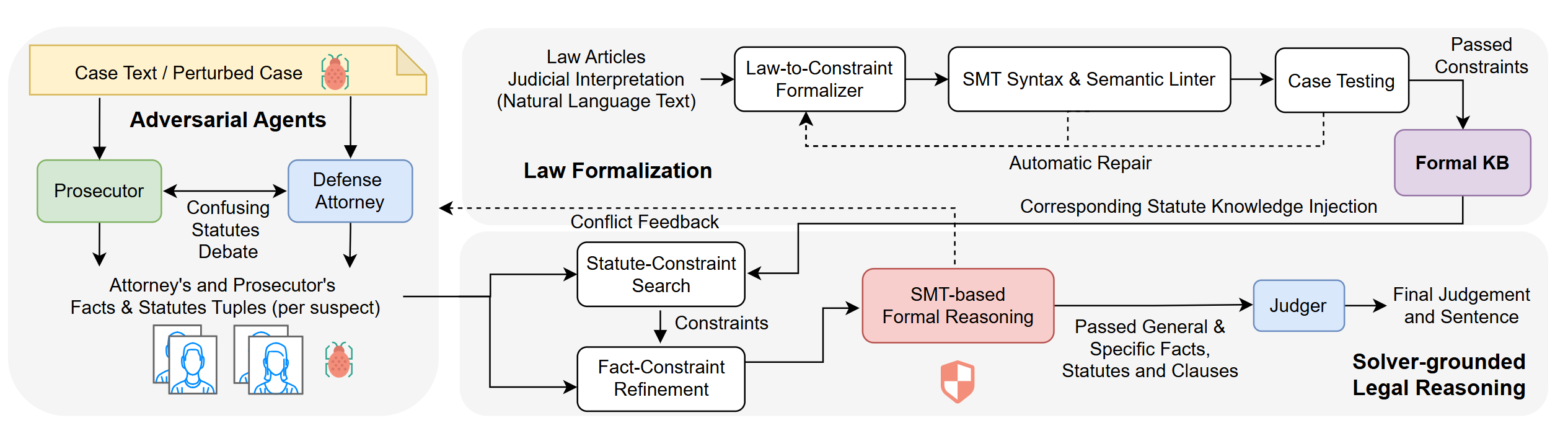}
    \caption{Overview of \method.
    (\textbf{Top}) \emph{Law Formalization}: statutes and judicial interpretations are translated into SMT-checkable legal constraints.
    (\textbf{Left}) \emph{Adversarial Agents}: prosecutor and defense agents independently extract facts and candidate statutes from the same case narrative.
    (\textbf{Bottom}) \emph{Solver-grounded Legal Reasoning}: encode the extracted facts and candidate statutes into a unified constraint set. The SMT solver checks statutory applicability, detects inconsistencies, and the judge returns a formally grounded judgment.}
    \label{fig:framework}
\end{figure*}

As shown in \autoref{fig:framework}, \method{} combines LLM-based legal interpretation with the formal reasoning. 
Given a case \(x\), the system outputs a judgment
$
y = \langle \text{statute}, \text{clause}, \text{penalty}, \text{explanation} \rangle,
$
where the explanation consists of checked legal conditions and supporting case facts. 
The key idea is to separate \emph{legal proposal} from \emph{legal check}: LLM agents identify potentially relevant facts and statutes, while the SMT solver determines whether those statutes can actually be applied under a formal legal knowledge base.
The detailed formalization is in Appendix\ref{app:formalization}.

\subsection{Law Formalization}
\label{subsec:formalization}
We first construct a formal legal knowledge base \(\mathcal{K}\) from statutes and judicial interpretations with the aid of LLMs. 
Each legal rule is represented as a conditional constraint.
The conditions specify when a statute or clause applies, and the legal effect specifies the resulting offense classification, liability, or penalty range.
Concretely, each statute article has an article-level guard, and each clause has a clause-level guard. 
The article guard checks whether the case falls within the general scope of a statute. 
The clause guard checks fine-grained legal requirements such as the actor, conduct, intent, harm, causation, protected legal interest, aggravating factors, mitigating factors, or statutory exceptions. 
A simplified rule is:
$
\textsc{ArticleGuard}_i \land \textsc{ClauseGuard}_{i,j}
\Rightarrow
\textsc{Penalty}_{i,j}.
$
The formal knowledge base is automatically generated from natural-language legal materials and then validated before use. 
We apply three validation steps. 
First, syntactic checking ensures that all generated rules are well-formed. 
Second, semantic checking detects contradictory, vacuous, or overly broad rules. 
Third, case-level testing checks whether known cases activate the expected statute and clause guards. 
Only validated constraints are included in \(\mathcal{K}\).

\subsection{Adversarial Agents}
\label{subsec:legal-llm}

Given the case narrative \(x\), \method{} uses two role-differentiated LLM agents: a prosecutor agent and a defense agent. 
Both agents read the same case, but they extract legal information from different argumentative perspectives. 
The prosecutor agent focuses on facts supporting liability, while the defense agent focuses on missing elements, exceptions, and alternative interpretations.

Each agent outputs a structured argument
$
\langle \mathcal{F}, \mathcal{L} \rangle .
$
The fact set \(\mathcal{F}\) contains suspect-centric facts extracted from the case narrative. 
Each fact is typed by a legally meaningful element, including actor, conduct, mental state, protected interest, object, amount, harm, consequence, causation, or exception, and is linked to a supporting text span. 
This grounding mitigates the risk of unsupported or hallucinated facts.
The candidate law set \(\mathcal{L}\) contains plausible statutes and clauses proposed by the agents. 
These proposals are not treated as final legal conclusions. 
Instead, we organize statutes into similarity clusters and prompt adversarial agents to debate confusing provisions within the same cluster. 
The SMT solver in the next step determines which proposed rules are actually applicable by checking them against the extracted facts and formalized legal constraints.

\subsection{Solver-Centered Adjudication}
\label{sec:solver_reasoning}

Given the fact set \(\mathcal{F}\), candidate statutes \(\mathcal{L}\), and the formal legal knowledge base \(\mathcal{K}\), \method{} does not encode all information at once. 
Instead, it first searches \(\mathcal{K}\) for the formal constraints corresponding to the proposed statutes:
$
\mathcal{C}_{\mathcal{L}} = \textsc{Search}(\mathcal{L}, \mathcal{K}).
$
These constraints specify the statutory elements, applicability conditions, exceptions, and penalty rules of each candidate article and clause.

Next, \method{} refines the extracted facts with respect to the retrieved constraints:
$
\mathcal{C}_{\mathcal{F}} = \textsc{Refine}(\mathcal{F}, \mathcal{C}_{\mathcal{L}}).
$
Each suspect-centric fact is matched to a required legal element, such as actor, conduct, mental state, protected interest, object, amount, harm, causation, consequence, or exception. 
The refined facts and statute constraints are then encoded into an SMT formula:
$
\Phi = \textsc{Encode}(\mathcal{C}_{\mathcal{F}}, \mathcal{C}_{\mathcal{L}}).
$
The SMT solver Z3 \cite{z3} checks the satisfiability of article guards, clause guards, and exception guards. 
Unsatisfied articles or clauses are discarded, while satisfied guards determine the applicable offense classification and penalty range. 
When multiple provisions are satisfiable, priority rules in \(\mathcal{K}\), including statutory hierarchy, specificity, exception handling, and penalty ordering, select the final applicable rule. 
Irrelevant facts are filtered out from formal reasoning.
Therefore, the final judgment is produced by solver-verified legal applicability rather than direct LLM generation.
If prosecutor and defense agents produce conflicting facts or incompatible statute proposals, the solver returns an unsat core that identifies the conflicting facts, missing elements, or incompatible guards. 
This conflict feedback triggers fact re-grounding or constraint repair. 
The reasoning loop continues until \(\Phi\) becomes satisfiable or all candidate statutes are rejected.

\section{Experiments}
We design the following five research questions. RQ1--RQ2 measure standard prediction performance and ablation effects, while RQ3--RQ5 test whether predictions remain legally grounded under our legal-relevance-sensitive evaluation.

\paragraph{Research Questions.}
\begin{itemize}
    \item \textbf{RQ1}: How accurately does \method{} predict applicable statutes and sentencing outcomes compared with legal and general LLM baselines?
    \item \textbf{RQ2}: Which components of \method{} contribute most to statute prediction?
    \item \textbf{RQ3}: Can \method{} make legally appropriate prediction changes under should-change perturbations?
    \item \textbf{RQ4}: Can \method{} remain stable under should-not-change perturbations?
    \item \textbf{RQ5}: What types of errors does \method{} make, especially in confusing statute cases?
\end{itemize}

\paragraph{Datasets.}
We use three datasets. 
LeCaRDv2~\cite{li2023lecardv2} contains 55,192 cases, with an average case-fact length of 889.17 Chinese characters and 4.53 applicable statutes per case, and supports case-level statute and sentence evaluation. 
LEEC~\cite{xue2024leec} provides suspect-level annotations for multi-defendant cases; we use 9,470 suspect-level instances, with an average fact length of 654.49 Chinese characters and 5.25 applicable statutes per instance. 
We further construct an 8,000-case controlled perturbation set from criminal-law fact patterns, with an average fact length of 134.89 Chinese characters and 1.65 statutes per case. 
The artifact is available on the anonymous website\cite{LexGuardweb}.

\paragraph{Implementation Details.}
We use the z3 \cite{z3} as the formal reasoner.
RQ3--RQ5 instantiate the legal-relevance-sensitive axes in Table~\ref{tab:legal_relevance_eval} through factual perturbations, prompt-injection families, and confusing-statute clusters, with all model outputs normalized to statute identifiers before scoring. 
Our legal-relevance-sensitive evaluation also follows the work J\&H~\cite{hu2025jh}. 
The baseline LLM-J\&H-CoT keeps the same output schema but uses a reasoning-augmented prompt, and LLM-J\&H-Few-shot uses example-calibrated demonstrations under the same output schema. 
Unless otherwise specified, \method{} uses GPT-5.2 as the base LLM for agent generation and final judgment, and GPT-5.2 is also used as the direct model baseline. 
Details appear in Appendix~\ref{app:settings} and Appendix~\ref{app:metrics}. 

\subsection{RQ1: Accuracy on Legal Prediction}
Table~\ref{tab:rq1_all} reports provision prediction results on LeCaRDv2 and LEEC. 
Across both datasets and provision granularities, \method{} consistently achieves the best F1 score. 
The gains mainly come from higher precision, indicating that solver-grounded verification helps suppress irrelevant statute predictions rather than merely increasing recall. 
This pattern is particularly clear on LeCaRDv2, where general-purpose LLMs often over-predict statutes and obtain moderate recall but low precision. 
It indicates that many baseline models fail not because they cannot retrieve any relevant statute, but because they cannot reliably reject similar yet inapplicable provisions.
On LEEC, the gap becomes larger under suspect-level evaluation, suggesting that provision prediction becomes substantially harder when the model must decompose case facts by individual defendants.

Table~\ref{tab:sentencing_combined} further evaluates downstream sentencing and suspect-level extraction. 
Providing golden statutes generally reduces sentencing error, confirming that statute misidentification is a major source of downstream punishment error. 
\method{} still achieves the lowest RMSE in both the w/o-golden and w/-golden settings on LeCaRDv2 and LEEC, while maintaining competitive or best legal validity. 
Overall, these results suggest that \method{} provides a more reliable reasoning pipeline: it selects fewer irrelevant provisions, produces legally valid outputs, and transfers these gains to downstream sentencing.

\begin{table*}[t]
\centering
\footnotesize
\setlength{\tabcolsep}{5pt}
\renewcommand{\arraystretch}{1.18}
\caption{Provision Prediction Performance on LeCaRDv2 and LEEC (\%).}
\label{tab:rq1_all}
\begin{tabular}{lcccccccccccc}
\toprule
 & \multicolumn{6}{c}{LeCaRDv2} & \multicolumn{6}{c}{LEEC (Suspect-Level)} \\
\cmidrule(lr){2-7} \cmidrule(lr){8-13}
 & \multicolumn{3}{c}{General} & \multicolumn{3}{c}{Specific}
 & \multicolumn{3}{c}{General} & \multicolumn{3}{c}{Specific} \\
\cmidrule(lr){2-4} \cmidrule(lr){5-7}
\cmidrule(lr){8-10} \cmidrule(lr){11-13}
Model
& P$\uparrow$ & R$\uparrow$ & F1$\uparrow$ & P$\uparrow$ & R$\uparrow$ & F1$\uparrow$
& P$\uparrow$ & R$\uparrow$ & F1$\uparrow$ & P$\uparrow$ & R$\uparrow$ & F1$\uparrow$ \\
\midrule
LexiLaw
& 8.96 & 26.27 & 13.36 & 47.76 & 49.91 & 48.81
& 1.03 & 0.21 & 0.34 & 1.89 & 3.09 & 2.23 \\
DISC-LawLLM
& 66.67 & 2.13 & 4.12 & 64.22 & 75.27 & 69.31
& 0.00 & 0.00 & 0.00 & 1.56 & 1.25 & 1.04 \\
GPT o4-mini
& 21.54 & 35.00 & 26.67 & 69.00 & 21.00 & 32.00
& 34.80 & 23.24 & 25.52 & 65.93 & 63.37 & 62.82 \\
GPT-4o
& 12.00 & 36.00 & 18.00 & 67.00 & 73.00 & 70.00
& 34.00 & 23.27 & 24.85 & 67.15 & 66.58 & 65.22 \\
Claude 4 Sonnet
& 32.79 & 26.67 & 29.41 & 64.00 & 75.00 & 69.00
& 36.85 & 24.05 & 26.05 & 65.96 & 63.63 & 63.73 \\
DeepSeek v3
& 10.25 & 44.84 & 16.88 & 60.19 & \textbf{82.77} & 69.70
& 32.16 & 22.95 & 23.29 & 64.72 & 63.49 & 62.50 \\
GPT-5.2
& 23.47 & \textbf{46.25} & 31.14 & 70.71 & 81.05 & 75.53
& 52.01 & 37.38 & 39.08 & 79.73 & 76.06 & 76.46 \\
\textbf{\method{} (Ours)}
& \textbf{34.00} & 45.95 & \textbf{39.08}
& \textbf{81.03} & 77.05 & \textbf{78.99}
& \textbf{64.05} & \textbf{40.77} & \textbf{45.51}
& \textbf{82.35} & \textbf{76.71} & \textbf{78.13} \\
\bottomrule
\end{tabular}
\end{table*}

\begin{table*}[t]
\centering
\footnotesize
\setlength{\tabcolsep}{5pt}
\renewcommand{\arraystretch}{1.18}
\caption{Sentencing error, legal validity, and suspect-level performance with or without golden statutes. RMSE denotes sentencing root mean square error in months. Valid denotes the proportion of solver-verified legally valid predictions. SusF1 denotes suspect-level F1 score on LEEC.}
\label{tab:sentencing_combined}
\begin{tabular}{lccccccccc}
\toprule
\textbf{Model}
& \multicolumn{4}{c}{\textbf{LeCaRDv2}}
& \multicolumn{5}{c}{\textbf{LEEC}} \\
\cmidrule(lr){2-5} \cmidrule(lr){6-10}
& \multicolumn{2}{c}{w/o Golden}
& \multicolumn{2}{c}{w/ Golden}
& \multicolumn{3}{c}{w/o Golden}
& \multicolumn{2}{c}{w/ Golden} \\
\cmidrule(lr){2-3} \cmidrule(lr){4-5} \cmidrule(lr){6-8} \cmidrule(lr){9-10}
& RMSE$\downarrow$ & Valid$\uparrow$
& RMSE$\downarrow$ & Valid$\uparrow$
& RMSE$\downarrow$ & Valid$\uparrow$ & SusF1$\uparrow$
& RMSE$\downarrow$ & Valid$\uparrow$ \\
\midrule
LexiLaw & 39.09 & 92.13 & 28.51 & 90.00 & 25.89 & 97.00 & 9.62 & 31.75 & 96.00 \\
DISC-LawLLM & 52.45 & 83.00 & 37.14 & 90.00 & -- & -- & 13.87 & -- & -- \\
GPT o4-mini & 33.20 & 92.00 & 28.03 & 94.00 & 34.94 & 91.00 & 78.68 & 38.46 & 96.00 \\
GPT-4o & 32.84 & 94.00 & 22.68 & 95.00 & 41.66 & 95.60 & 80.25 & 50.41 & 95.50 \\
Claude 4 Sonnet & 26.25 & 93.00 & 26.20 & 94.00 & 36.92 & 94.00 & 89.97 & 28.40 & \textbf{98.00} \\
DeepSeek v3 & 26.73 & 64.30 & 15.42 & 93.00 & 27.67 & 97.00 & 66.48 & 22.61 & 97.00 \\
GPT 5.2 & 14.54 & 94.20 & 13.87 & 94.00 & 28.48 & 95.20 & 97.80 & 23.90 & 95.00 \\
\textbf{\method{} (Ours)} & \textbf{12.72} & \textbf{94.60} & \textbf{9.98} & \textbf{96.20} & \textbf{23.04} & \textbf{97.20} & \textbf{98.80} & \textbf{20.95} & 96.77 \\
\bottomrule
\end{tabular}
\end{table*}

\subsection{RQ2: Ablation Study}

Table~\ref{tab:ablation_results} reports the ablation study on LeCaRDv2. 
The full \method{} achieves the best F1 for both general and specific provisions, indicating that the three components are complementary. 
Removing the Z3 reasoner leads to the largest degradation in statute prediction. 
This shows that symbolic consistency checking is crucial for filtering legally invalid or internally inconsistent statute candidates. 
Removing the debating module also substantially reduces both G-F1 and S-F1, confirming that fine-grained competition among confusable provisions is necessary for accurate statute selection. 
Removing the attorney module mainly weakens adversarial coverage: recall drops for both general and specific provisions, suggesting that the prosecutor--defense decomposition helps expose missing candidate statutes. 

\begin{table*}[t]
\centering
\footnotesize
\setlength{\tabcolsep}{4pt}
\renewcommand{\arraystretch}{1.18}
\caption{Ablation study results with different components removed. G-P, G-R, and G-F1 denote precision, recall, and F1 for general-provision prediction; S-P, S-R, and S-F1 denote precision, recall, and F1 for specific-provision prediction. All methods are based on GPT-5.2.}
\label{tab:ablation_results}
\begin{tabular}{lcccccc}
\toprule
\textbf{Model Variant} & \textbf{G-P$\uparrow$} & \textbf{G-R$\uparrow$} & \textbf{G-F1$\uparrow$} & \textbf{S-P$\uparrow$} & \textbf{S-R$\uparrow$} & \textbf{S-F1$\uparrow$} \\
\midrule
-- Attorney & \textbf{36.02} & 39.91 & 37.87 & 74.22 & 65.53 & 69.60 \\
-- Z3 Reasoner & 29.29 & 22.74 & 25.60 & 66.70 & 56.14 & 60.97 \\
-- Statutes Debating & 31.75 & 26.87 & 29.10 & 74.30 & 61.79 & 67.47 \\
Raw LLM & 22.63 & \textbf{47.69} & 30.69 & 65.45 & 75.00 & 69.90 \\
\textbf{\method{} (Ours)} & 34.31 & 46.07 & \textbf{39.33} & \textbf{80.98} & \textbf{77.04} & \textbf{78.96} \\
\bottomrule
\end{tabular}
\end{table*}

\subsection{RQ3: Should-Change to Perturbations}
RQ3 evaluates whether models can make legally appropriate prediction changes when the case facts are modified in ways that should alter statutory applicability. 
Table~\ref{tab:rq3_rq4_main} reports performance under should-change factual perturbations. 
The results show that \method{} handles should-change perturbations more reliably than all baselines. 
This indicates that \method{} is better at recognizing when a factual modification changes the applicable statutory conditions and at updating its statute prediction in the legally expected direction.
The improvement is especially important because should-change perturbations test legal sensitivity rather than mere stability.

\begin{table*}[t]
\centering
\footnotesize
\setlength{\tabcolsep}{4pt}
\renewcommand{\arraystretch}{1.18}
\caption{RQ3 and RQ4 results. RQ3: Overall denotes counterfactual statute-set score, Align. denotes change-alignment score, Sta. denotes statute correctness score, and Bias denotes perturbation-factor bias magnitude. RQ4: ASR denotes attack success rate, CRR denotes clean-correct retention rate, Inv. denotes prediction invariance, and F1 denotes attack-aware prediction F1. All methods are based on GPT-5.2.}
\label{tab:rq3_rq4_main}
\begin{tabular}{lcccccccc}
\toprule
\textbf{Model}
& \multicolumn{4}{c}{\textbf{RQ3}}
& \multicolumn{4}{c}{\textbf{RQ4}} \\
\cmidrule(lr){2-5} \cmidrule(lr){6-9}
& Overall$\uparrow$ & Align.$\uparrow$ & Sta.$\uparrow$ & Bias$\downarrow$
& ASR$\downarrow$ & CRR$\uparrow$ & Inv.$\uparrow$ & F1$\uparrow$ \\
\midrule
\method{} 
& \textbf{0.8101} & \textbf{0.8353} & \textbf{0.7702} & \textbf{0.0242}
& \textbf{0.4988} & \textbf{0.1571} & \textbf{0.2602} & \textbf{0.5588} \\
LLM 
& 0.7235 & 0.7434 & 0.6911 & 0.0904
& 0.5625 & 0.0082 & 0.0573 & 0.4013 \\
LLM-J\&H-CoT 
& 0.7341 & 0.7503 & 0.7076 & 0.0996
& 0.5417 & 0.0100 & 0.0401 & 0.4209 \\
LLM-J\&H-Few-shot 
& 0.7360 & 0.7359 & 0.7361 & 0.0920
& 0.5357 & 0.0238 & 0.0735 & 0.4493 \\
\bottomrule
\end{tabular}
\end{table*}

\subsection{RQ4: Should-Not-Change Perturbations}
RQ4 evaluates whether models can maintain legally correct predictions under adversarial perturbations that should not change the applicable statute. 
As shown in Table~\ref{tab:rq3_rq4_main}, \method{} achieves the strong robustness across all metrics. 
It is less likely to be misled by injected legal distractions, better preserves originally correct predictions, and produces more stable statute sets under attack. The improvement in attack-aware F1 further shows that \method{} does not merely keep predictions unchanged, but maintains legally accurate decisions when facing should-not-change perturbations.

The J\&H prompting baselines provide only limited protection. CoT and few-shot prompting improve robustness slightly over the vanilla LLM, but their clean-correct retention and invariance remain weak, suggesting that prompt-level reasoning guidance cannot reliably prevent the model from following misleading but legally irrelevant cues. In contrast, \method{} verifies extracted facts and candidate statutes against formalized legal constraints, helping distinguish legally operative conditions from adversarial noise. These results indicate that robustness to should-not-change perturbations requires more than prompt engineering; it benefits from solver-grounded verification that enforces consistency with the governing legal rule.

\subsection{RQ5: Distinguishing Confusing Statutes}

RQ5 analyzes model errors in legally confusing statute clusters. 
Table~\ref{tab:rq5_main} reports both cluster-level exactness and error types. 
The results show that \method{} is substantially better at fully recovering applicable statutes when a confusing cluster is relevant, achieving 88.71\% positive exactness compared with 58.57\% for the base LLM. 
It also achieves the highest macro cluster exactness, indicating more balanced performance across different similar-statute clusters. 
By contrast, few-shot prompting reduces omission rate, but increases wrong similar-statute selection, suggesting that demonstrations make the model more willing to activate cluster statutes without reliably resolving statutory boundaries. 
\method{} reduces both error types, lowering omission and wrong selection. 
This suggests that reverse verification helps the system not only select the applicable member of a confusing cluster, but also reject legally similar yet inapplicable alternatives.
The false-activation diagnostic and cluster-level breakdowns are provided in Appendix~\ref{app:error_analysis}.

\begin{table*}[t]
\centering
\footnotesize
\setlength{\tabcolsep}{5pt}
\renewcommand{\arraystretch}{1.18}
\caption{RQ5 error analysis. Pos. denotes exact selection when the gold confusing-statute cluster is non-empty; Macro denotes macro-averaged cluster exactness across statute clusters; Omit denotes gold-statute omission; Wrong denotes wrong similar-statute selection. All methods are based on GPT-5.2.}
\label{tab:rq5_main}
\begin{tabular}{lcccc}
\toprule
\textbf{Model} & Pos.$\uparrow$ & Macro$\uparrow$ & Omit$\downarrow$ & Wrong$\downarrow$ \\
\midrule
\method{} & \textbf{88.71\%} & \textbf{79.17\%} & \textbf{9.68\%} & \textbf{1.61\%} \\
LLM & 58.57\% & 59.47\% & 39.39\% & 6.39\% \\
LLM-J\&H-CoT & 55.75\% & 56.60\% & 42.20\% & 6.39\% \\
LLM-J\&H-Few-shot & 62.53\% & 61.73\% & 35.29\% & 7.54\% \\
\bottomrule
\end{tabular}
\end{table*}


\subsection{Limitations}
Despite its promising results, our framework still faces three main challenges. 
First, its formalization quality is constrained by the accuracy of LLM outputs, so errors in identifying actors or conditions can propagate throughout the reasoning pipeline. 
Second, our current formalization system is limited to statutory rules, leaving the extension to case law, open-textured norms, and evolving jurisprudence as future work. 
Third, the system assumes deterministic rule parsing, which prevents it from fully capturing legal provisions that deliberately incorporate normative ambiguity.
In addition, the framework introduces moderate computational overhead due to its multi-stage LLM-based reasoning process.
Detailed costs are shown in the appendix (\autoref{tab:cost_analysis}).

\section{Related Work}

\paragraph{Domain-specific Legal LLMs.}
General-purpose LLMs often mishandle legal terminology and citation style.  
\textsc{ChatLaw} couples a knowledge-graph-enhanced mixture-of-experts backbone with a multi-agent pipeline that mirrors law-firm SOPs, outperforming GPT-4 on LawBench and national bar exams \cite{cui2023chatlaw}.  
\textsc{Lawyer GPT} shows that lightweight domain pre-training plus retrieval boosts statute-matching and consultation accuracy while remaining compute-efficient \cite{yao2024lawyergpt}.  
Gao et al. \cite{gao2024enhancing} further show that careful construction of high-quality synthetic query–candidate pairs can markedly improve legal-case retrieval.
\textsc{Agents on the Bench} simulates a collegial bench to improve judgment quality through deliberative voting \cite{jiang2024agentsbench}, whereas \textsc{AgentCourt} evolves adversarial lawyer agents via long-horizon self-play, yielding measurable skill gains \cite{chen2024agentcourt}. 
These systems, however, lack guarantees on logical soundness.  
We retain the agent metaphor and introduce the legal-relevance-sensitive evaluation and the \emph{neural–symbolic} pipeline.
Our work goes further by compiling the extracted norms into \emph{executable} formal reasoning.

\paragraph{Legal LLM Evaluation and Benchmarks.}
\textsc{LegalBench} constructs a collaboratively designed benchmark of 162 legal-reasoning tasks and evaluates both open-source and commercial LLMs across different forms of legal reasoning~\cite{guha2023legalbench}.
\textsc{LawBench} focuses on Chinese legal tasks and organizes 20 tasks according to Bloom's cognitive taxonomy, providing a more jurisdiction-specific view of legal knowledge memorization, understanding, and application~\cite{fei2024lawbench}.
\textsc{LEXTREME} extends legal NLP evaluation to multilingual and multi-task settings, covering 11 datasets across 24 languages and showing that legal-domain evaluation remains challenging even for strong language models~\cite{niklaus2023lextreme}.
More recently, \textsc{LegalAgentBench} evaluates LLM agents in Chinese legal scenarios with real-world corpora, external legal tools, and progress-based intermediate metrics~\cite{li2025legalagentbench}.
Our evaluation is complementary: instead of measuring only task-level accuracy, we stress-test whether legal predictions change for legally justified reasons.
Unlike J\&H~\cite{hu2025j}, which only studies label-preserving knowledge-injection robustness, our framework additionally evaluates should-change legal sensitivity. 
We introduce relevance-sensitive evaluation and provide a solver-grounded mechanism.

\paragraph{Neural Symbolic Methods.}  
Early work such as Logic Tensor Networks \cite{badreddine2020ltn} embeds many-valued fuzzy logic into differentiable architectures, while DeepProbLog \cite{manhaeve2018deepproblog} integrates probabilistic logic programming with neural predicates.  
More recent studies add large-scale LLM components: NS-LCR learns explicit law- and case-level rules for explainable case retrieval \cite{sun2024nslcr}; Logic-LM introduces structured prompting plus theorem-prover feedback to enforce logical soundness \cite{sadowski2025logiclm}; and Kant et al. combine neuro-symbolic reasoning with contract analysis to improve coverage decisions \cite{kant2025robust}.  
These results show that logic grounding enhances transparency and robustness, which we pursue by marrying adversarial agents with a symbolic solver for trustworthy legal AI.

\section{Conclusion}

In this paper, we frame trustworthy legal AI as a problem of \emph{legal relevance sensitivity}: models should remain stable under legally irrelevant changes and respond appropriately to legally material ones. 
We show that existing LLMs often fail to distinguish relevant legal changes from irrelevant perturbations. 
To address these failures, we propose \method, a solver-grounded multi-stage reasoning framework that anchors legal decisions in explicit statutory conditions and verifiable reasoning. 
Our results suggest that trustworthy legal AI requires not only high accuracy, but also stable, fair, and legally grounded sensitivity to what truly matters under law.

\newpage

\bibliographystyle{plain}
\bibliography{reference}

\clearpage
\appendix
\captionsetup[table]{skip=6pt}
\section{Cost Statistics}
\begin{table}[htbp]
\centering
\footnotesize
\setlength{\tabcolsep}{5pt}
\renewcommand{\arraystretch}{1.18}
\caption{Average cost per case of \method{}.}
\label{tab:cost_analysis}
\begin{tabular}{p{0.58\columnwidth} p{0.28\columnwidth}}
\toprule
\textbf{Metric} & \textbf{Average} \\
\midrule
LLM calls per case & 10.33 \\
Runtime (sec) & 107.36 \\
Input tokens & 13{,}049 \\
Output tokens & 4{,}217 \\
Cost per case (USD) & 0.0819 \\
\bottomrule
\end{tabular}
\end{table}
Table~\ref{tab:cost_analysis} shows that \method{} incurs a moderate computational overhead per case. On average, each case requires 10.33 LLM calls and 107.36 seconds of runtime, with a total cost of only \$0.0819. This suggests that, despite employing a multi-stage agent-and-solver pipeline, \method{} remains practically affordable for legal reasoning tasks. The token statistics further indicate that most of the cost comes from processing relatively long legal inputs, while the overall per-case monetary expense remains low.

\section{Details of Statutory Formalization}
\label{app:formalization}

This appendix provides the formal specification used to translate statutory provisions into
solver-checkable constraints. At a high level, each statutory clause is represented as an implication
from an article-level applicability guard and a clause-level condition guard to a legally admissible
penalty range:
\[
    \textsc{ArticleGuard}(a,x) \wedge \textsc{ClauseGuard}(c,x)
    \Rightarrow \textsc{Penalty}(c,x),
\]
where \(a\) denotes an article, \(c\) denotes a clause under the article, and \(x\) denotes a suspect-centric
case representation. We make this template explicit below.

\subsection{Typed Case Representation}

We represent each case as a tuple of typed legal facts:
\[
x = \langle s, v, A, R, M, Q, E \rangle ,
\]
where \(s\) is the suspect, \(v\) is the victim or protected legal interest, \(A\) is the set of acts,
\(R\) is the set of harmful results, \(M\) is the mental state, \(Q\) is the set of qualifying circumstances,
and \(E\) is the set of legally irrelevant or extra-legal attributes.

We use the following typed predicates:
\[
\begin{aligned}
&\textsc{Actor}(s), \quad \textsc{Victim}(v), \quad \textsc{Act}(s,a), \quad \textsc{Result}(r),\\
&\textsc{Causes}(a,r), \quad \textsc{MentalState}(s,m), \quad \textsc{ProtectedInterest}(v,i),\\
&\textsc{Amount}(r,n), \quad \textsc{Severity}(r,\ell), \quad \textsc{Qualifier}(s,q),\\
&\textsc{ExtraLegal}(e), \quad \textsc{HasExtraLegalAttr}(x,e).
\end{aligned}
\]

The domains of core symbols are typed as:
\[
\begin{aligned}
&s \in \mathcal{S}, \quad v \in \mathcal{V}, \quad a \in \mathcal{A}, \quad r \in \mathcal{R},\\
&m \in \{\textsf{Intentional}, \textsf{Negligent}, \textsf{Knowing}, \textsf{Unknown}\},\\
&\ell \in \{\textsf{Minor}, \textsf{Serious}, \textsf{EspeciallySerious}\},\\
&q \in \mathcal{Q}, \quad e \in \mathcal{E}.
\end{aligned}
\]

For implementation, these predicates are encoded as Boolean or integer SMT variables. For example,
a case-level fact such as ``the suspect intentionally caused serious injury'' is encoded as:
\[
\textsc{MentalState}(s,\textsf{Intentional}) = \top,
\quad
\textsc{Severity}(r,\textsf{Serious}) = \top,
\quad
\exists a \, [\textsc{Act}(s,a) \wedge \textsc{Causes}(a,r)].
\]

\subsection{Article-Level Applicability Guard}

Each article \(a\) is first associated with an article-level guard. This guard specifies the general legal
domain in which the article may apply. It prevents the solver from considering provisions that are
categorically irrelevant to the case.

For an article \(A_k\), the article guard is defined as:
\[
\textsc{ArticleGuard}_{k}(x)
=
\textsc{SubjectGuard}_{k}(x)
\wedge
\textsc{ObjectGuard}_{k}(x)
\wedge
\textsc{ConductGuard}_{k}(x).
\]

More explicitly:
\[
\begin{aligned}
\textsc{SubjectGuard}_{k}(x)
    &:= \bigwedge_{p \in \mathcal{P}^{subj}_{k}} p(s),\\
\textsc{ObjectGuard}_{k}(x)
    &:= \bigvee_{i \in \mathcal{I}_{k}} \textsc{ProtectedInterest}(v,i),\\
\textsc{ConductGuard}_{k}(x)
    &:= \exists a \in A \; \textsc{ActType}_{k}(a).
\end{aligned}
\]

Here, \(\mathcal{P}^{subj}_{k}\) is the set of subject requirements, such as being a natural person,
state functionary, employee, or responsible person; \(\mathcal{I}_{k}\) is the set of protected legal
interests covered by the article; and \(\textsc{ActType}_{k}\) denotes the conduct type regulated by
the article.

Thus, an article is activated only if:
\[
\textsc{ApplicableArticle}_{k}(x)
\Leftrightarrow
\textsc{ArticleGuard}_{k}(x).
\]

\subsection{Clause-Level Guard}

A statutory article may contain multiple clauses corresponding to different factual thresholds,
aggravating circumstances, mitigating circumstances, or penalty brackets. For each clause \(C_{k,j}\)
under article \(A_k\), we define:
\[
\textsc{ClauseGuard}_{k,j}(x)
=
\textsc{ElementGuard}_{k,j}(x)
\wedge
\textsc{ThresholdGuard}_{k,j}(x)
\wedge
\textsc{ExceptionGuard}_{k,j}(x).
\]

The element guard checks whether the necessary legal elements are satisfied:
\[
\textsc{ElementGuard}_{k,j}(x)
=
\textsc{MentalGuard}_{k,j}(x)
\wedge
\textsc{ResultGuard}_{k,j}(x)
\wedge
\textsc{CausationGuard}_{k,j}(x).
\]

The mental-state guard is:
\[
\textsc{MentalGuard}_{k,j}(x)
=
\bigvee_{m \in \mathcal{M}_{k,j}}
\textsc{MentalState}(s,m),
\]
where \(\mathcal{M}_{k,j}\) is the set of legally admissible mental states for the clause.

The result guard is:
\[
\textsc{ResultGuard}_{k,j}(x)
=
\bigvee_{r \in R}
\left[
    \textsc{ResultType}_{k,j}(r)
    \wedge
    \textsc{Severity}_{k,j}(r)
\right].
\]

The causation guard is:
\[
\textsc{CausationGuard}_{k,j}(x)
=
\exists a \in A, r \in R \;
[
    \textsc{Act}(s,a)
    \wedge
    \textsc{Result}(r)
    \wedge
    \textsc{Causes}(a,r)
].
\]

The threshold guard encodes quantitative requirements, such as amount, number of victims, loss
value, injury level, or repetition:
\[
\textsc{ThresholdGuard}_{k,j}(x)
=
\bigwedge_{t \in \mathcal{T}_{k,j}}
\textsc{SatisfyThreshold}(x,t).
\]

For an amount-based threshold \(t=(r,n,\bowtie,\theta)\), this becomes:
\[
\textsc{SatisfyThreshold}(x,t)
\Leftrightarrow
\textsc{Amount}(r,n) \wedge n \bowtie \theta .
\]

The exception guard excludes clauses that are blocked by legally recognized exceptions:
\[
\textsc{ExceptionGuard}_{k,j}(x)
=
\neg
\bigvee_{e \in \mathcal{X}_{k,j}}
\textsc{Exception}_{e}(x).
\]

Therefore, a clause is admissible iff:
\[
\textsc{AdmissibleClause}_{k,j}(x)
\Leftrightarrow
\textsc{ArticleGuard}_{k}(x)
\wedge
\textsc{ClauseGuard}_{k,j}(x).
\]

\subsection{Penalty Encoding}

Each admissible clause maps to a legally permitted penalty interval. We encode the sentence as an
integer variable:
\[
y \in \mathbb{Z}_{\ge 0},
\]
where \(y\) denotes the imprisonment term in months. Special punishments such as life imprisonment
or death penalty can be represented by reserved symbols:
\[
y \in \mathbb{Z}_{\ge 0} \cup \{\textsf{Life}, \textsf{Death}\}.
\]

For ordinary fixed-term imprisonment, each clause \(C_{k,j}\) defines a penalty interval:
\[
\textsc{Penalty}_{k,j}(y)
\Leftrightarrow
L_{k,j} \le y \le U_{k,j},
\]
where \(L_{k,j}\) and \(U_{k,j}\) are the lower and upper statutory bounds.

The complete clause implication is:
\[
\forall x,y \; .
\left[
    \textsc{ArticleGuard}_{k}(x)
    \wedge
    \textsc{ClauseGuard}_{k,j}(x)
\right]
\Rightarrow
\textsc{Penalty}_{k,j}(y).
\]

Equivalently:
\[
\forall x,y \; .
\textsc{AdmissibleClause}_{k,j}(x)
\Rightarrow
L_{k,j} \le y \le U_{k,j}.
\]

\subsection{Aggravating and Mitigating Circumstances}

Some statutory clauses do not directly define a separate offense, but modify the penalty range.
We represent aggravating and mitigating factors as monotonic transformations over the base interval.

Let:
\[
B_{k,j}(x) = [L_{k,j}, U_{k,j}]
\]
be the base penalty interval. Let \(\mathcal{G}(x)\) be the set of aggravating factors and
\(\mathcal{M}(x)\) the set of mitigating factors.

An aggravating factor \(g\) is encoded as:
\[
\textsc{Aggravating}_{g}(x)
\Rightarrow
y \ge L_{k,j} + \Delta^{-}_{g},
\]
where \(\Delta^{-}_{g}\) raises the effective lower bound.

A mitigating factor \(m\) is encoded as:
\[
\textsc{Mitigating}_{m}(x)
\Rightarrow
y \le U_{k,j} - \Delta^{+}_{m},
\]
where \(\Delta^{+}_{m}\) lowers the effective upper bound.

Thus, the adjusted penalty interval is:
\[
\textsc{AdjustedPenalty}_{k,j}(x,y)
\Leftrightarrow
L'_{k,j}(x) \le y \le U'_{k,j}(x),
\]
where:
\[
L'_{k,j}(x)
=
L_{k,j}
+
\sum_{g \in \mathcal{G}(x)}
\Delta^{-}_{g},
\]
and:
\[
U'_{k,j}(x)
=
U_{k,j}
-
\sum_{m \in \mathcal{M}(x)}
\Delta^{+}_{m}.
\]

To avoid inconsistent intervals, we require:
\[
L'_{k,j}(x) \le U'_{k,j}(x).
\]

If this constraint is unsatisfiable, the solver rejects the corresponding statutory interpretation.

\subsection{Invariance to Legally Irrelevant Attributes}

To enforce legal relevance, extra-legal attributes are explicitly excluded from article and clause
guards. For any extra-legal attribute \(e \in \mathcal{E}\), we require:
\[
e \notin
\textsc{Vars}
\left(
    \textsc{ArticleGuard}_{k}
    \cup
    \textsc{ClauseGuard}_{k,j}
    \cup
    \textsc{Penalty}_{k,j}
\right).
\]

Equivalently, for two cases \(x\) and \(x'\) that differ only in extra-legal attributes:
\[
x \equiv_{\neg \mathcal{E}} x'
\Rightarrow
\textsc{AdmissibleClause}_{k,j}(x)
=
\textsc{AdmissibleClause}_{k,j}(x').
\]

The predicted statutory set should therefore remain invariant:
\[
x \equiv_{\neg \mathcal{E}} x'
\Rightarrow
\mathcal{C}^{*}(x) = \mathcal{C}^{*}(x'),
\]
where:
\[
\mathcal{C}^{*}(x)
=
\{C_{k,j} \mid \textsc{AdmissibleClause}_{k,j}(x)\}.
\]

This constraint is used to distinguish legally irrelevant perturbations from legally material changes.

\subsection{Solver Objective}

Given extracted facts \(F_x\) and formalized statutory knowledge base \(\mathcal{K}\), the solver searches
for admissible clauses and penalty assignments:
\[
\mathcal{K} \cup F_x \models
\textsc{AdmissibleClause}_{k,j}(x)
\wedge
\textsc{AdjustedPenalty}_{k,j}(x,y).
\]

The set of solver-supported clauses is:
\[
\mathcal{C}^{*}(x)
=
\left\{
C_{k,j}
\mid
\mathcal{K} \cup F_x
\models
\textsc{AdmissibleClause}_{k,j}(x)
\right\}.
\]

The final penalty set is:
\[
\mathcal{Y}^{*}(x)
=
\left\{
y
\mid
\exists C_{k,j} \in \mathcal{C}^{*}(x),
\mathcal{K} \cup F_x
\models
\textsc{AdjustedPenalty}_{k,j}(x,y)
\right\}.
\]

If multiple clauses are admissible, we apply statutory priority rules. Specifically, more specific
clauses dominate more general clauses:
\[
C_{k,j_1} \succ C_{k,j_2}
\quad \text{iff} \quad
\textsc{ClauseGuard}_{k,j_1}(x)
\Rightarrow
\textsc{ClauseGuard}_{k,j_2}(x)
\]
and not conversely. The selected clause set is:
\[
\widehat{\mathcal{C}}(x)
=
\left\{
C \in \mathcal{C}^{*}(x)
\mid
\nexists C' \in \mathcal{C}^{*}(x), C' \succ C
\right\}.
\]

The final sentence is then selected from the legally valid interval:
\[
\widehat{y}
\in
\bigcup_{C_{k,j} \in \widehat{\mathcal{C}}(x)}
[L'_{k,j}(x), U'_{k,j}(x)].
\]

\subsection{Example Schema}

For illustration, consider a generic criminal article \(A_k\) with two clauses. Clause \(C_{k,1}\)
covers ordinary circumstances, while clause \(C_{k,2}\) covers serious circumstances.

The article guard is:
\[
\textsc{ArticleGuard}_{k}(x)
=
\textsc{Actor}(s)
\wedge
\exists a \in A \; \textsc{RegulatedAct}_{k}(a)
\wedge
\exists r \in R \; \textsc{ProtectedResult}_{k}(r).
\]

The ordinary clause is:
\[
\begin{aligned}
\textsc{ClauseGuard}_{k,1}(x)
=
&\textsc{MentalState}(s,\textsf{Intentional})
\wedge
\exists a,r \; [
    \textsc{Act}(s,a)
    \wedge
    \textsc{Causes}(a,r)
    \wedge
    \textsc{Severity}(r,\textsf{Minor})
]
\\
&\wedge
\neg \textsc{EspeciallySeriousCircumstance}(x).
\end{aligned}
\]

Its penalty range is:
\[
\textsc{Penalty}_{k,1}(y)
\Leftrightarrow
0 \le y \le 36.
\]

The serious clause is:
\[
\begin{aligned}
\textsc{ClauseGuard}_{k,2}(x)
=
&\textsc{MentalState}(s,\textsf{Intentional})
\wedge
\exists a,r \; [
    \textsc{Act}(s,a)
    \wedge
    \textsc{Causes}(a,r)
    \wedge
    \textsc{Severity}(r,\textsf{Serious})
]
\\
&\vee
\textsc{EspeciallySeriousCircumstance}(x).
\end{aligned}
\]

Its penalty range is:
\[
\textsc{Penalty}_{k,2}(y)
\Leftrightarrow
36 < y \le 120.
\]

The full formalization is therefore:
\[
\begin{aligned}
&
\textsc{ArticleGuard}_{k}(x)
\wedge
\textsc{ClauseGuard}_{k,1}(x)
\Rightarrow
0 \le y \le 36,
\\
&
\textsc{ArticleGuard}_{k}(x)
\wedge
\textsc{ClauseGuard}_{k,2}(x)
\Rightarrow
36 < y \le 120.
\end{aligned}
\]

This example shows how the compact template
\(\textsc{ArticleGuard} \wedge \textsc{ClauseGuard} \Rightarrow \textsc{Penalty}\)
is instantiated as a typed, clause-specific, solver-checkable representation. In our framework, LLMs
are used only to propose candidate facts and candidate statutory mappings, while the final admissible
clauses and penalty ranges must be validated by the solver against the formalized statutory knowledge
base.

\section{Dataset Details}
\label{app:dataset_details}

\subsection*{Dataset 1: LeCaRDv2 Subset Dataset}
The structure of the original candidate set of the LeCaRDv2 dataset is shown in Table~\ref{tab:lecard_raw}.

\begin{table}[htbp]
\centering
\footnotesize
\setlength{\tabcolsep}{5pt}
\renewcommand{\arraystretch}{1.18}
\caption{Structure of a raw LeCaRDv2 case.}
\label{tab:lecard_raw}
\begin{tabular}{p{0.25\columnwidth} p{0.62\columnwidth}}
\toprule
\textbf{Field} & \textbf{Example} \\
\midrule
filename & 3554630.json \\
fact & Tianjin Baodi District Court criminal judgment... \\
article & [264, 64, 65, 67] \\
result & Natural language court ruling \\
\bottomrule
\end{tabular}
\end{table}

We use the \texttt{fact} field as the model input. The \texttt{result} field is parsed to extract the sentencing result in months, saved as \texttt{true\_sentence\_months} for automated evaluation. We also split the \texttt{article} field into \texttt{true\_general\_articles} and \texttt{true\_specific\_articles} in accordance with Chinese Criminal Law to allow separate evaluation. The processed structure is shown in Table~\ref{tab:lecard_used}.

\begin{table}[htbp]
\centering
\footnotesize
\setlength{\tabcolsep}{5pt}
\renewcommand{\arraystretch}{1.18}
\caption{Structure of a processed LeCaRDv2 case.}
\label{tab:lecard_used}
\begin{tabular}{p{0.38\columnwidth} p{0.50\columnwidth}}
\toprule
\textbf{Field} & \textbf{Example} \\
\midrule
filename & 3554630.json \\
fact & Full case text including... \\
article & [264, 64, 65, 67] \\
true\_sentence\_months & 44 \\
true\_general\_articles & [64, 65, 67] \\
true\_specific\_articles & [264] \\
\bottomrule
\end{tabular}
\end{table}

\subsection*{Dataset 2: LEEC Suspect-Level Dataset}
We construct a suspect-level evaluation dataset based on the publicly released LEEC corpus. We first apply rule-based regular expressions to recover structured fields, and then parse sentencing decisions into suspect-level labels, including each suspect's charge and applicable article list. The processed suspect-level format used in our experiments is shown in Table~\ref{tab:leec_used}.

\begin{table}[htbp]
\centering
\footnotesize
\setlength{\tabcolsep}{5pt}
\renewcommand{\arraystretch}{1.18}
\caption{Structure of the processed LEEC suspect-level format.}
\label{tab:leec_used}
\begin{tabular}{p{0.22\columnwidth} p{0.66\columnwidth}}
\toprule
\textbf{Field} & \textbf{Example} \\
\midrule
pid & 2 \\
qw & Full judicial judgment text including court, defendants, facts, and ruling \\
fact & Narrative of criminal facts involving multiple defendants \\
reason & Court reasoning and legal analysis \\
result & Natural language sentencing decisions for each defendant \\
charge & \{Suspect A: Traffic Accident Crime; Suspect B: Harboring Crime; Suspect C: Harboring Crime\} \\
article & \{Suspect A: [133, 310, 67, 72]; Suspect B: [133, 310, 67, 72]; Suspect C: [133, 310, 67, 72]\} \\
\bottomrule
\end{tabular}
\end{table}

\subsection*{Dataset 3: Controlled Perturbation Dataset for RQ3}
The RQ3 dataset evaluates whether a legal reasoning system makes the expected prediction change when case facts are modified in legally material ways. It is stored as \texttt{perturbed\_cases\_nips.json} and contains 8,000 paired criminal-law cases. Each pair consists of a base case and a perturbed case, together with structured metadata describing the perturbation rule, perturbation category, label-change status, and before/after statute labels. The reported statistics are computed from the perturbed case facts and suspect-level statute annotations. This design supports should-change sensitivity tests and related label-preserving diagnostics.

\begin{table}[htbp]
\centering
\footnotesize
\setlength{\tabcolsep}{5pt}
\renewcommand{\arraystretch}{1.18}
\caption{Structure of the RQ3 controlled perturbation dataset.}
\label{tab:rq3_perturb_struct}
\resizebox{\columnwidth}{!}{%
\begin{tabular}{p{0.22\columnwidth} p{0.30\columnwidth} p{0.38\columnwidth}}
\toprule
\textbf{Field} & \textbf{JSON key} & \textbf{Description} \\
\midrule
ID & \texttt{perturbation\_id} & Unique identifier of the perturbation instance, e.g., \texttt{C1\_P1}. \\
Original case & \texttt{original\_case\_id} & Identifier of the corresponding base case. \\
Template & \texttt{template\_type} & Crime-template family used to instantiate the case, such as bribery, drug crime, traffic accident, or property crime. \\
Rules & \texttt{perturbation\_rules} & Atomic perturbation rules applied to the base case, such as adding surrender facts or changing role-related facts. \\
Categories & \texttt{perturbation\_categories} & Legal-relevance categories targeted by the perturbation, including general-provision, role, subjective-state, and statutory-element edits. \\
Label change & \texttt{changed\_label} & Boolean annotation indicating whether the applicable statute set should change after perturbation. \\
Legal effect & \texttt{label\_effect} & Explanation of the expected legal effect, including before/after statute labels for affected suspects. \\
Cases & \texttt{base\_case}; \texttt{perturbed\_case} & Structured original and perturbed cases, each containing case facts and suspect-level statute annotations. \\
\bottomrule
\end{tabular}%
}
\end{table}

\begin{table}[htbp]
\centering
\footnotesize
\setlength{\tabcolsep}{5pt}
\renewcommand{\arraystretch}{1.18}
\caption{Example record from the RQ3 perturbation dataset.}
\label{tab:rq3_perturb_example}
\resizebox{\columnwidth}{!}{%
\begin{tabular}{p{0.20\columnwidth} p{0.34\columnwidth} p{0.38\columnwidth}}
\toprule
\textbf{Field} & \textbf{Value} & \textbf{Description} \\
\midrule
ID & \texttt{C1\_P1} & Unique perturbation instance. \\
Template & \texttt{bribery\_tripartite} & Crime-template family used to instantiate the case. \\
Rules & \texttt{add\_intermediary}; \texttt{add\_for\_benefit}; \texttt{add\_self\_surrender} & Atomic edits applied to the base case. \\
Categories & \texttt{subjective\_and\_role}; \texttt{general} & Legal-relevance categories targeted by the edits. \\
Label change & \texttt{true} & The applicable statute set is expected to change. \\
Legal effect & Article~67 is introduced for the affected suspect. & Role- and benefit-related edits test bribery-element sensitivity. \\
Statistics & 8,000 cases; 134.89 chars; 1.65 statutes. & Dataset-level count, average fact length, and average statute count. \\
\bottomrule
\end{tabular}%
}
\end{table}

\section{Experiment Setup Details}\label{app:settings}

\subsection*{Hardware Environment}
All experiments were conducted on a dedicated research server equipped with two NVIDIA RTX 6000 Ada GPUs, each with 48GB VRAM, under CUDA 12.4 and driver version 550.144.03. The software environment used Anaconda-managed Python environments for model execution and evaluation.

\subsection*{Evaluation Protocol}
For RQ1, statute prediction is formulated as multi-label classification. We report precision, recall, and F1 for general provisions and specific provisions on LeCaRDv2 and LEEC. Sentencing is evaluated with sentencing error (SE), RMSE, and legal validity, with and without golden statutes. LEEC additionally reports suspect extraction F1 because each case may contain multiple defendants. RQ2 disables one major component at a time and evaluates the resulting change in general precision, recall, F1, specific precision, recall, F1.

For RQ3, every instance contains an original case, a perturbed case, perturbation metadata, and before/after statute labels. We score predictions against the post-perturbation gold statute set, focusing on whether the model changes its prediction when the perturbation changes statutory applicability. For RQ4, adversarial user content is appended to the legal input while the gold legal label remains unchanged, following the should-not-change prompt-injection setting described in the main text. RQ5 provides a confusing-statute cluster and requires the model to select only applicable articles inside that cluster, or abstain when the cluster is irrelevant.

\subsection*{Baselines and Prompt Variants}
We compare \method{} against direct LLM inference and specialized legal LLMs. GPT-5.2, GPT-4o, GPT o4-mini, DeepSeek v3, and Claude 4 Sonnet are evaluated through API calls. DISC-LawLLM~\cite{yue2023disc} and LexiLaw~\cite{LexiLaw} are legal-domain baselines. GPT-5.2-J\&H-CoT keeps the same output schema as GPT-5.2 but adds an explicit legal reasoning scaffold for rule identification, fact-to-element mapping, elimination of inapplicable alternatives, and final prediction. GPT-5.2-J\&H-Few-shot uses the same evaluation scripts and API configuration as the corresponding J\&H-CoT runs, but calibrates the prompt with legal input-output demonstrations rather than a procedural reasoning scaffold.

\subsection*{Implementation Details}
All systems are evaluated with deterministic scripts and normalized statute identifiers. \method{} applies statute extraction, fact extraction, competitive-article refinement, and solver-centric verification before final prediction. For baseline outputs that are generated in natural language, we use the same structured extraction prompts to recover article identifiers and sentencing outcomes before computing metrics.

\section{Additional Experimental Results and Robustness Protocols}
\label{app:additional_experiment_details}

\subsection{Metric Definitions}
\label{metrics}
\label{app:metrics}

Table~\ref{tab:app_metric_definitions} defines the metrics used in Table~\ref{tab:legal_relevance_eval} and in the experimental results. Predictions and labels are normalized to statute identifiers before evaluation. Higher values are better unless marked with $\downarrow$.
The framework table uses the same metric names as the experiment tables. Abbreviated headers such as Align., Sta., Inv., ASR, CRR, Pos., Macro, Omit, and Wrong refer to Change Alignment, Statute Correctness, Invariance, Attack Success Rate, Clean-correct Retention Rate, Positive Exactness, Macro Exactness, Gold Omission, and Wrong Similar Selection, respectively.

\begin{table*}[tbp]
\centering
\footnotesize
\setlength{\tabcolsep}{5pt}
\renewcommand{\arraystretch}{1.18}
\caption{Metric definitions for the evaluation framework and experiments.}
\label{tab:app_metric_definitions}
\begin{tabular}{p{3.4cm} p{9.4cm}}
\toprule
Metric & Definition \\
\midrule
\multicolumn{2}{l}{\textit{Statute prediction and sentencing metrics}} \\
P / R / F1 $\uparrow$ & Precision, recall, and F1 for predicted general or specific statute sets. \\
SE $\downarrow$ & Sentencing error in months. \\
RMSE $\downarrow$ & Root mean squared sentencing error in months. \\
Valid Ratio $\uparrow$ & Fraction of sentencing outputs that satisfy statutory range and consistency checks. \\
Suspect F1 $\uparrow$ & F1 between predicted and gold suspect sets in multi-defendant LEEC cases. \\
G-P / G-R / G-F1 $\uparrow$ & Precision, recall, and F1 for general-provision prediction under each ablation variant. \\
S-P / S-R / S-F1 $\uparrow$ & Precision, recall, and F1 for specific-provision prediction under each ablation variant. \\

\midrule
\multicolumn{2}{l}{\textit{Legal-relevance and counterfactual robustness metrics}} \\
Overall Score $\uparrow$ & Average post-perturbation statute matching score over general and specific provisions. \\
Change Alignment $\uparrow$ & Average score on label-changing perturbations, where legal labels are expected to change. \\
Statute Correctness / Sta.$\uparrow$ &
Exact statute-set correctness after a should-change perturbation, measuring whether the post-perturbation prediction exactly matches the updated gold statute set. \\
Invariance $\uparrow$ & Average score on label-preserving perturbations, where legal labels should remain stable. \\
Bias Magnitude $\downarrow$ & Weighted performance shift across perturbation-factor groups relative to form/noise perturbations. \\

\midrule
\multicolumn{2}{l}{\textit{Attack robustness metrics}} \\
ASR $\downarrow$ & Fraction of clean-correct paired cases that become incorrect after attack. \\
CRR $\uparrow$ & Fraction of paired cases that are correct before attack and remain correct after attack. \\
Attack Invariance $\uparrow$ & Fraction of paired cases whose extracted legal prediction remains unchanged under attack. \\
Attack Precision / Recall / F1 $\uparrow$ & Statute-set precision, recall, and F1 under adversarial perturbations.
Attack-aware F1 denotes the aggregated F1 across evaluated attacked cases. \\

\midrule
\multicolumn{2}{l}{\textit{Confusing-statute discrimination metrics}} \\
Positive Exactness $\uparrow$ & Exact cluster-article match when the confusing cluster contains at least one gold article. \\
Macro Exactness $\uparrow$ & Macro-averaged exact cluster-article match across confusing-statute clusters. \\
Gold Omission $\downarrow$ & Fraction of positive-cluster cases where at least one gold cluster article is missed. \\
Wrong Similar Sel. $\downarrow$ & Fraction of positive-cluster cases selecting a non-gold article from the same confusing-statute cluster. \\
False Activation $\downarrow$ & Fraction of negative-cluster cases where the model incorrectly selects a cluster article. \\
\bottomrule
\end{tabular}
\end{table*}

\section{Breakdown Results for Robustness Experiments}
\label{app:error_analysis}

The main text reports aggregate robustness results. Tables~\ref{tab:rq3_category_breakdown}--\ref{tab:rq5_cluster_breakdown}
provide compact breakdowns by perturbation category, prompt-injection template,
and confusing-statute cluster. These breakdowns are intended to show whether the
observed trends are concentrated in a narrow subset of cases or persist across
legally distinct stress conditions.

\begin{table}[htbp]
\centering
\footnotesize
\caption{RQ3 results by perturbation category for \method{}. Scores are percentages; score columns are averaged over successfully returned cases.}
\label{tab:rq3_category_breakdown}
\begin{tabular}{lrrrrrr}
\toprule
\textbf{Category} & \textbf{Inst.} & \textbf{Changed} & \textbf{Success} & \textbf{Overall} & \textbf{General} & \textbf{Specific} \\
\midrule
Behavior path & 98 & 69.4 & 93.9 & 82.4 & 83.9 & 80.9 \\
Expression & 65 & 41.5 & 81.5 & 83.2 & 88.2 & 78.1 \\
General provision & 49 & 75.5 & 85.7 & 83.2 & 85.5 & 81.0 \\
Noise & 29 & 55.2 & 93.1 & 79.8 & 83.6 & 75.9 \\
Subjective state/role & 24 & 29.2 & 83.3 & 71.2 & 91.2 & 51.2 \\
Amount & 18 & 50.0 & 77.8 & 78.0 & 93.5 & 62.5 \\
Consequence & 13 & 92.3 & 100.0 & 73.7 & 88.5 & 59.0 \\
\bottomrule
\end{tabular}%
\end{table}

\begin{table}[htbp]
\centering
\footnotesize
\caption{RQ4 prompt-injection breakdown for \method{}. Clean and attack accuracies, ASR, invariance, and F1 are reported as percentages.}
\label{tab:rq4_attack_breakdown}
\resizebox{\columnwidth}{!}{%
\begin{tabular}{p{1.4cm} p{2.8cm}rrrrrr}
\toprule
\textbf{Dataset} & \textbf{Attack template} & \textbf{Pairs} & \textbf{Clean Acc.} & \textbf{Attack Acc.} & \textbf{ASR} & \textbf{Inv.} & \textbf{F1} \\
\midrule
LeCaRDv2 & Fabricated authority & 38 & 10.5 & 5.3 & 50.0 & 28.9 & 28.8 \\
LeCaRDv2 & Format mimicking & 36 & 8.3 & 2.8 & 100.0 & 30.6 & 30.6 \\
LeCaRDv2 & Role hijacking & 35 & 11.4 & 5.7 & 75.0 & 25.7 & 29.3 \\
LeCaRDv2 & Verdict forcing & 36 & 11.1 & 5.6 & 50.0 & 27.8 & 30.0 \\
LEEC & Fabricated authority & 32 & 7.7 & 2.6 & 66.7 & 35.9 & 31.9 \\
LEEC & Format mimicking & 30 & 5.4 & 2.7 & 50.0 & 37.8 & 29.6 \\
LEEC & Role hijacking & 30 & 8.1 & 5.4 & 66.7 & 29.7 & 35.0 \\
LEEC & Verdict forcing & 31 & 7.9 & 2.6 & 66.7 & 23.7 & 33.4 \\
\bottomrule
\end{tabular}%
}
\end{table}

\begin{table*}[tbp]
\centering
\footnotesize
\setlength{\tabcolsep}{5pt}
\renewcommand{\arraystretch}{1.18}
\caption{RQ5 cluster-level breakdown for \method{} on non-perfect confusing-statute clusters. All rate columns are percentages.}
\label{tab:rq5_cluster_breakdown}
\begin{tabular}{p{3.0cm} p{3.2cm}rrrrr}
\toprule
\textbf{Cluster} & \textbf{Articles} & \textbf{Eval.} & \textbf{Hit} & \textbf{Exact} & \textbf{Miss} & \textbf{Wrong} \\
\midrule
Aviation & 121, 122 & 4 & 25.0 & 25.0 & 75.0 & 0.0 \\
Bribery by public officer & 164, 385, 386 & 4 & 100.0 & 75.0 & 0.0 & 25.0 \\
Drug storage & 363, 364 & 4 & 0.0 & 0.0 & 100.0 & 0.0 \\
Contract/financial fraud & 176, 192--194, 224--226, 266 & 4 & 75.0 & 75.0 & 25.0 & 0.0 \\
Marriage and family & 258, 2601 & 3 & 66.7 & 66.7 & 33.3 & 0.0 \\
Occupation-property & 163, 271, 272 & 4 & 25.0 & 25.0 & 75.0 & 0.0 \\
Food/drug product & 140, 144, 197, 198 & 3 & 33.3 & 33.3 & 66.7 & 0.0 \\
Safety accident & 133, 135 & 4 & 0.0 & 0.0 & 100.0 & 0.0 \\
Smuggling & 151, 152 & 4 & 0.0 & 0.0 & 100.0 & 0.0 \\
Traffic safety & 117, 120 & 4 & 25.0 & 25.0 & 75.0 & 0.0 \\
Weapons/explosives & 125, 128 & 4 & 50.0 & 50.0 & 50.0 & 0.0 \\
\bottomrule
\end{tabular}
\end{table*}

\subsection{Robustness Benchmark Construction}
\paragraph{Should-change factual perturbations for RQ3.}
The factual perturbation benchmark is built from paired original and perturbed criminal-law fact patterns. Each perturbed instance records whether the legal label should change, the perturbation category, and the before/after statutory labels. In the main text, RQ3 emphasizes should-change cases, where a model should update its statute prediction when a factual modification changes statutory applicability.

\paragraph{Should-not-change prompt injection for RQ4.}
The prompt-injection benchmark evaluates whether models preserve legal judgment under adversarial user content. We instantiate four attack families: fabricated authority, verdict forcing, role hijacking, and format mimicking. The attacks are appended to legal inputs while the gold legal label remains unchanged, so degradation reflects susceptibility to external instructions rather than a genuine change in case facts.

\paragraph{Confusing statutes.}
The confusing-statute benchmark groups legally similar provisions into clusters. Positive cases contain at least one applicable statute from the cluster, while negative cases contain no applicable statute from that cluster. This setup evaluates both omission errors on relevant clusters and false activation errors on irrelevant clusters.

\subsection{Reasoning-Augmented GPT-5.2 Baseline}
\label{app:cot_baseline}

We construct GPT-5.2-J\&H-CoT by adding a task-level reasoning scaffold to the baseline prompt while keeping the input cases and output schema fixed. The scaffold is organized into four blocks. First, an issue-framing block asks the model to identify the legally relevant question raised by the case. Second, a rule-identification block asks for candidate statutory provisions and the legal elements that must be satisfied. Third, an element-mapping block requires the model to align concrete case facts with each statutory element and mark unsupported elements. Fourth, a contrastive-exclusion block asks the model to compare legally similar alternatives and remove provisions whose elements are not established. Only after these blocks does the prompt request the final structured prediction.

The scaffold is specialized for each robustness task. For RQ3, it requires an explicit comparison between the base and perturbed facts and asks whether the perturbation changes a legally material condition before predicting the post-perturbation statutes. For RQ4, it separates adjudicative facts from adversarial instructions and directs the model to ground the decision only in legally relevant case content. For RQ5, it performs cluster-level contrastive screening: each candidate provision in the confusing-statute cluster is checked against its elements, and the model must either select the supported article(s) or abstain from activating the cluster.

This construction does not provide gold statutes, perturbation labels, solver states, or intermediate outputs from \method{}. GPT-5.2-J\&H-CoT is therefore a controlled reasoning-augmented prompting baseline: it tests whether an explicit legal-reasoning scaffold can improve a strong LLM under the same structured-output interface used for direct GPT-5.2.

GPT-5.2-J\&H-Few-shot uses the same structured output interface and evaluation scripts, but replaces the procedural reasoning scaffold with task-specific input-output demonstrations. For RQ3, examples illustrate label-preserving and label-changing factual edits; for RQ4, examples show that adversarial appended instructions should be separated from adjudicative facts; for RQ5, examples demonstrate exact selection inside a confusing-statute cluster and abstention when the cluster is irrelevant. The demonstrations do not reveal test labels or solver states, and are used only to calibrate the model's response format and decision pattern.

\section{Prompt Details}

This section presents the prompt templates used in each major module of our main workflow, covering statute selection, fact extraction, structured schema alignment, formal KB-based reasoning, and final decision explanation. All prompts are designed to elicit interpretable and structured outputs from LLMs, facilitating integration with downstream modules and evaluation logic.

\subsection*{Statute Selection and Fact Extraction Prompts}

\subsubsection*{Attorney Agent Prompts}

\paragraph{Attorney StatuteSelectorAgent Prompt}\mbox{}\\
\textit{Purpose: Select applicable legal provisions from the Criminal Law of the People's Republic of China for \textbf{defense} purposes.}

\begin{quote}
You are a legal expert in the field of Chinese Criminal Law. Based on the following case description, determine the potentially applicable legal provisions from the Criminal Law of the People's Republic of China. Divide your output into ``General Provisions'' and ``Specific Provisions''. Your objective is to \textbf{support} the prosecution’s position in this case.

\textbf{Case Description:} \{case\_text\}

Please return strictly in the following JSON format:  
\{  
\hspace{2em} "general\_articles": [article numbers, e.g., 17],\\
\hspace{2em} "specific\_articles": [article numbers, e.g., 234]  
\}
\end{quote}

\paragraph{Attorney FactExtractorAgent Prompt}\mbox{}\\
\textit{Purpose: Extract comprehensive case facts in natural Chinese language from \textbf{defense} purposes.}

\begin{quote}
You are a legal fact extraction expert. Please extract factual information as comprehensively as possible from the following case description. Return the facts in well-organized, semantically clear natural Chinese language. Your goal is to \textbf{defend} the case.

\textbf{Case Description:} \{case\_text\}

Do not return JSON. Present the facts using paragraph or bullet-list format.
\end{quote}

\vspace{1em}

\subsubsection*{Prosecutor Agent Prompts}

\paragraph{Prosecutor StatuteSelectorAgent Prompt}\mbox{}\\
\textit{Purpose: Select applicable legal provisions from the Criminal Law of the People's Republic of China from the \textbf{prosecution} perspective.}

\begin{quote}
You are a legal expert in the field of Chinese Criminal Law. Based on the following case description, determine the potentially applicable legal provisions from the Criminal Law of the People's Republic of China. Divide your output into ``General Provisions'' and ``Specific Provisions''. Your objective is to \textbf{accuse} in this case.

\textbf{Case Description:} \{case\_text\}

Please return strictly in the following JSON format:  
\{  
\hspace{2em} "general\_articles": [article numbers, e.g., 17],\\
\hspace{2em} "specific\_articles": [article numbers, e.g., 234]  
\}
\end{quote}

\paragraph{Prosecutor FactExtractorAgent Prompt}\mbox{}\\
\textit{Purpose: Extract comprehensive case facts in natural Chinese language for \textbf{accusation} purposes.}

\begin{quote}
You are a legal fact extraction expert. Please extract as detailed factual information as possible from the following case description. Return the facts in well-organized, semantically clear natural Chinese language. Your goal is to \textbf{accuse} in this case.

\textbf{Case Description:} \{case\_text\}

Do not return JSON. Present the facts using paragraph or bullet-list format.
\end{quote}

\subsection*{Structured Schema Extraction Prompt}

\textit{Purpose: To extract structured factual elements from raw LLM outputs according to the predefined schema in our Formal KB, including both General and Specific fields. This enables Statute-Fact alignment in reverse verification.}

\begin{quote}
The following two analysts provide descriptions of the case facts. Based on their descriptions, please extract structured legal elements:

\textbf{Analyst 1:} \{fact\_extractor\_output\}

\textbf{Analyst 2:} \{fact\_reviewer\_text\}

Please extract:

1. \textbf{General Provision Fields} (Only return those marked as true):
\{GENERAL\_FIELD\_LIST\}
Example output:
\{
  "age\_under\_18": true,
  "truthful\_confession\_of\_crime": true
\}

2. \textbf{Specific Provision Fields} (select values only from the given options):
- subject: \{SPECIFIC\_FIELDS['subject']\}
- action: \{SPECIFIC\_FIELDS['action']\}
- object: \{SPECIFIC\_FIELDS['object']\}
- intent: \{SPECIFIC\_FIELDS['intent']\}

Output format:
\{
  "general\_facts": \{
    "age\_under\_18": true,
    ...
  \},
  "specific\_facts": \{
    "subject": [...],
    "action": [...],
    ...
  \}
\}
\end{quote}

\subsection*{Law-Specific Fact Slicing Prompt}

\textit{Purpose: Given a specific article and its Formal KB-defined value ranges, this prompt guides the LLM to generate a fully structured fact slice suitable for formal reasoning.}

\begin{quote}
You are a legal expert familiar with the Criminal Law of the People's Republic of China. Based on the following case description and the value range of Article \{article\}, generate a valid JSON input that matches the requirements of this article.

Requirements:
1. Output must be strictly in structured JSON format and include only the following fields: \{list(range\_data.keys())\}.
2. Each field must be assigned a single value from the provided range.
3. The ``Age'' field can be an integer or null; all other fields are strings.
4. If information is insufficient, choose the most reasonable default.
5. Severity-related fields such as ``Condition'' and ``Quantity'' should be chosen conservatively and leniently.
6. Do not include code blocks, comments, or non-JSON content.

\textbf{Case Description:} \{case\_text\}

\textbf{Value Ranges:} \{range\_json\}

\textbf{Example Output:}
\{
  "Actor": "person",
  "Age": null,
  "Action": "selling",
  "Quantity": "small",
  "Condition": "none"
\}
\end{quote}

\subsection*{Final Judgment Explanation Prompt}

\textit{Purpose: Used by the Judge LLM to generate a human-readable final decision summary based on the structured results, without reanalyzing the original case. This ensures interpretability and transparency.}

\begin{quote}
You are a legal expert familiar with the Criminal Law of the People's Republic of China. Based on the following judgment results, write a final decision summary strictly in the specified format.

Requirements:
- Only and strictly use the statutes, charges, and penalties mentioned in the judgment result.
- Do not reanalyze the case description.
- Base your summary on the ``consequences'' and ``model\_details'' fields in the judgment.
- \{''.join(requirements)\}
- Output must strictly follow the given format and must not contain any extra commentary.

\textbf{Case Description (for reference only):} \{case\_text\}

\textbf{Judgment Result:} \{results\_json\}
\end{quote}

\section{Motivating Example Details}
\label{app:motivating_example_details}

This appendix provides a complete walkthrough of the motivating example discussed in the main text, illustrating how each module processes the case to produce a structured and legally grounded decision.
For completeness, we also include the two tables moved from the main text: (i) statute-specific KB examples used for auto-formalizing (Table~\ref{tab:article_kb}) and (ii) the pseudo-code and example execution of solver-based statutory sentencing for drug-related crimes (Table~\ref{tab:z3_reasoner}).

\subsection{Statute-Specific Knowledge Bases for Auto-formalizing}
\label{app:statute_kb_details}

This subsection presents representative statute-specific knowledge bases used in the structured fact extraction and auto-formalizing stage. These knowledge bases define the required legal fields and constraints for both general and specific provisions, enabling consistent schema-based fact extraction and subsequent statute-specific fact slicing.

\begin{table*}[tbp]
\centering
\footnotesize
\setlength{\tabcolsep}{5pt}
\renewcommand{\arraystretch}{1.18}
\caption{
Examples of law-specific knowledge bases for fine-grained fact extraction.
}
\label{tab:article_kb}
\begin{tabular}{p{1.6cm} p{6.2cm} p{6.4cm}}
\toprule
\textbf{Article} & \textbf{Relevant Fields (from Sentence KB)} & \textbf{Description and Legal Implication} \\
\midrule
347 (Drug Trafficking) &
\textit{Actor} (person, unit);  
\textit{Age} (integer, $\geq$12 years, default $\geq$18);  
\textit{Action} (smuggling, selling, transporting, manufacturing);  
\textit{DrugQuantity} (opium/heroin/meth thresholds in grams);  
\textit{Circumstance} (ringleader, armed protection, international trafficking);  
\textit{Condition} (none) &
Defines factors affecting sentencing for narcotics crimes, including actor type, age, type of drug activity, drug quantity thresholds, and aggravating circumstances (e.g., organized or armed operations). These parameters determine the severity level of Article 347 sentencing rules. \\
\midrule
65 (Recidivism) &
\textit{PriorSentenceType} (fixed-term, life, death penalty reprieve, other);  
\textit{NewCrimeSentenceType} (fixed-term, life, death penalty, other);  
\textit{TimeSincePriorSentence} (within 5 years, $>$5 years);  
\textit{TimeSinceParoleCompletion} (within 5 years, $>$5 years);  
\textit{CrimeIntent} (intentional, negligent);  
\textit{ParoleStatus} (on parole, not on parole);  
\textit{Condition} (none) &
Captures prior conviction details and their temporal relationship with the new offense, allowing the system to determine if the recidivism clause applies and whether sentence aggravation rules under Article 65 are triggered. \\
\bottomrule
\end{tabular}
\end{table*}

\subsection{Formal Solver-Based Statutory Reasoning}
\label{app:formal_reasoning_details}

This subsection details the formal reasoning process used for statute verification and sentencing determination.
We provide the pseudo-code of the Z3-based solver and a concrete execution trace for Article~347 (drug trafficking), illustrating how legal conditions are evaluated in severity order to derive a legally admissible sentencing clause.

\begin{table*}[tbp]
\centering
\footnotesize
\setlength{\tabcolsep}{5pt}
\renewcommand{\arraystretch}{1.18}
\caption{
Formal reasoning for statutory sentencing: pseudo-code and drug-crime example.
}
\label{tab:z3_reasoner}
\begin{tabular}{p{0.48\linewidth} p{0.48\linewidth}}
\toprule
\textbf{Algorithm} & \textbf{Example Execution: Article 347 (Drug Trafficking)} \\
\midrule
\textbf{Input:} Fact slice $f$, statute-specific rules $\Phi$ \newline
\textbf{Output:} Sentencing clause $C$ \newline
1: Initialize solver $S$, set $C=\emptyset$ \newline
2: Define valid actors, actions, drug quantities, circumstances \newline
3: Encode facts $f$ into solver $S$ \newline
4: For each rule $r_i \in \Phi$ (high $\rightarrow$ low severity): \newline
5:\quad Push solver state \newline
6:\quad Encode $r_i$ constraints:\newline
7:\quad\quad $r_1$: Large quantity or severe circumstance $\rightarrow$ 15y/life/death \newline
8:\quad\quad $r_2$: Medium quantity (10g-50g meth) $\rightarrow$ $\geq$7y \newline
9:\quad\quad $r_3$: Small quantity + serious circumstance $\rightarrow$ 3-7y \newline
10:\quad~ $r_4$: Small quantity + none circumstance $\rightarrow$ $\leq$3y \newline
11:\quad Check satisfiability of $r_i$ \newline
12:\quad If SAT and $C=\emptyset$: \newline
13:\quad\quad $C \leftarrow$ consequence linked to $r_i$ \newline
14:\quad\quad Pop and \textbf{break} \newline
15:\quad Pop solver state \newline
16: If $C=\emptyset$: $C=$ "No applicable clause" \newline
17: \textbf{return} $C$
&
1: Initialize solver with empty state \newline
2: Recognize Actor = person, Action = selling narcotics \newline
3: Encode facts (Actor, Action, DrugQuantity = heroin/meth $<$ 10g) \newline
4: Iterate over rules $r_1$ to $r_4$ in severity order \newline
5: Push solver state for $r_1$ \newline
6: Encode large quantity rule constraints \newline
7: Evaluate $r_1$ $\Rightarrow$ UNSAT (large quantity condition not met) \newline
8: Push solver state for $r_2$, encode medium quantity rule \newline
9: Evaluate $r_2$ $\Rightarrow$ UNSAT (medium quantity condition not met) \newline
10: Push solver state for $r_3$, encode small quantity + serious circumstance \newline
11: Evaluate $r_3$ $\Rightarrow$ UNSAT (no serious circumstance) \newline
12: Push solver state for $r_4$, encode small quantity + none circumstance \newline
13: Evaluate $r_4$ $\Rightarrow$ SAT (conditions satisfied) \newline
14: Set $C =$ "Fixed-term imprisonment $\leq$ 3 years", pop and break \newline
15: Discard other solver states \newline
16: Verify non-empty $C$, ready for return \newline
17: Return final sentencing clause under Article 347
\\
\bottomrule
\end{tabular}
\end{table*}

\subsection{Running Example Outputs}

This subsection consolidates the running example into the motivating-example appendix.
Each stage briefly states its role and then reports the corresponding output produced in the example.

\subsubsection*{Full Case Description (Model Input)}

The following is the full text of the original judgment, which serves as the input to the model:

\textit{Jiangxi Province Nanchang City Donghu District People's Court Criminal Judgment (2020) Gan 0102 Xingchu No. 347}

Procuratorate: Donghu District People's Procuratorate, Nanchang City.

Defendant: Gong Qiang, male, born February 10, 1974, Han ethnicity, junior high school education, employee of Xinchang Power Plant, registered in Donghu District, Nanchang City, Jiangxi Province.

Previous drug-related administrative detentions and compulsory detoxifications occurred between 2015 and 2019. Gong Qiang was criminally detained in November 2019 on suspicion of drug trafficking and formally arrested on November 19, 2019.

The prosecution accused that on August 30, 2019, at 13:32, the defendant received a WeChat message from Luo requesting to purchase drugs worth RMB 1,000. After receiving payment via WeChat, at approximately 17:00 the same day, Gong Qiang delivered 2 grams of methamphetamine to Luo. He was arrested on November 7, 2019, and confessed to the crime. Drug tests for both Gong and Luo returned positive for methamphetamine.

\subsubsection*{Outputs from Prosecutor Agent and Attorney Agent}

The role-differentiated legal agents propose candidate statutes and extract supporting facts from different legal perspectives.

\paragraph{Prosecutor Agent Outputs:}\mbox{}\\

Statute Selector output:\\ 
\texttt{\{general\_articles: [65, 67], specific\_articles: [347]\}}\\

Fact Extractor output:
\begin{itemize}
\item 1.1 Defendant: Gong Qiang
\item 1.2 Buyer: Luo (purchaser of the drugs)
\item 2.1 August 30, 2019, 13:32: Luo messages Gong via WeChat to purchase drugs
\item 2.2 Payment: RMB 1,000 via WeChat
\item 2.3 17:00: 2g methamphetamine handed over by Gong
\item 3.1 November 7: Gong surrenders at police station
\item 3.2 Gong confesses without resistance
\item 4.1 Urine tests conducted by police
\item 4.2 Tests returned meth-positive for both parties
\item 5.1 Gong previously convicted multiple times for similar crimes within 5 years
\item 6.1 Transaction conducted via mobile payment; highly concealed
\item 6.2 Gong fully cooperated with the investigation
\end{itemize}

\paragraph{Attorney Agent Outputs:}\mbox{}\\

Statute Selector 2 output:\\ 
\texttt{\{general\_articles: [67, 64], specific\_articles: [347]\}}\\

Fact Extractor 2 output:
\begin{itemize}
\item August 30, 2019, 13:32: Gong receives Luo's message
\item 17:00: Drug delivery (2g methamphetamine)
\item November 7: Gong voluntarily turns himself in
\item RMB 1,000 payment confirmed
\item Positive drug tests for both parties
\item Prior convictions within 5 years
\item Continued criminal activity post-incarceration
\end{itemize}

\subsubsection*{Autoformalize and Field Intersection Results}

The autoformalizer converts the extracted facts into statute-specific structured fields and intersects them with each candidate article's trigger conditions.

\textbf{General Provision Structured Facts Extracted:}
\begin{lstlisting}[language=json]
{
  "prior_sentence_served_or_pardoned": true,
  "reoffense_within_5_years": true,
  "recidivist_status": true,
  "voluntary_surrender_with_confession": true,
  "truthful_confession_of_crime": true,
  "provision_of_significant_clue": true,
  "multiple_crimes_before_sentencing": true,
  "illegal_proceeds_obtained": true
}
\end{lstlisting}

\textbf{Specific Provision Structured Facts Extracted:}
\begin{lstlisting}[language=json]
{
  "subject": ["person"],
  "action": ["traffic_narcotics"],
  "object": ["narcotics"],
  "intent": ["intentional"]
}
\end{lstlisting}

\textbf{Field Intersections with Trigger Conditions:}
\begin{lstlisting}[language=json]
Article 64 required: ["illegal_proceeds_obtained", ...]
Matched: {"illegal_proceeds_obtained": true}

Article 65 required: ["prior_sentence_served_or_pardoned", "reoffense_within_5_years", ...]
Matched: {"prior_sentence_served_or_pardoned": true, "reoffense_within_5_years": true}

Article 67 required: ["voluntary_surrender_with_confession", "truthful_confession_of_crime", ...]
Matched: {"voluntary_surrender_with_confession": true, "truthful_confession_of_crime": true}

Article 347 required: ["subject", "action", "object", "intent"]
Matched: {
  "subject": ["person"],
  "action": ["traffic_narcotics"],
  "object": ["narcotics"],
  "intent": ["intentional"]
}
\end{lstlisting}

\textbf{Final Structured Slices:}
\begin{lstlisting}[language=json]
{
  "general": [
    {
      "article": 64,
      "fields": {"illegal_proceeds_obtained": true}
    },
    {
      "article": 65,
      "fields": {
        "prior_sentence_served_or_pardoned": true,
        "reoffense_within_5_years": true
      }
    },
    {
      "article": 67,
      "fields": {
        "voluntary_surrender_with_confession": true,
        "truthful_confession_of_crime": true
      }
    }
  ],
  "specific": [
    {
      "article": 347,
      "fields": {
        "subject": ["person"],
        "action": ["traffic_narcotics"],
        "object": ["narcotics"],
        "intent": ["intentional"]
      }
    }
  ]
}
\end{lstlisting}

\subsubsection*{Verified Articles by Z3 Backward Verification}

The reverse-verification solver checks which candidate articles are legally triggerable from the structured fact slices.

Verified General Articles: [64, 65, 67]\\
Verified Specific Articles: [347]\\

\subsubsection*{Statute-specific Sentencing Outputs from Z3}

The statute-specific solver computes the legal consequences attached to the verified provisions.

\begin{lstlisting}[language=json, caption={Z3-based Law-specific Sentencing Outputs}]
{
  "Article 64": {
    "consequences": ["property shall be confiscated and turned over to the state treasury"],
    "model_details": {
      "Condition": "none",
      "Actor": "person",
      "PropertyType": "contraband"
    }
  },
  "Article 65": {
    "consequences": ["the offender is a recidivist and shall receive a heavier punishment under Article 65"],
    "model_details": {
      "PriorSentenceType": "fixed-term imprisonment",
      "CrimeIntent": "intentional",
      "ParoleStatus": "not on parole",
      "Condition": "none",
      "TimeSinceParoleCompletion": "within 5 years",
      "NewCrimeSentenceType": "fixed-term imprisonment",
      "TimeSincePriorSentence": "within 5 years"
    }
  },
  "Article 67": {
    "consequences": ["punishment may be mitigated"],
    "model_details": {
      "CompulsoryMeasure": "none",
      "Outcome": "none",
      "Actor": "person",
      "Action": "voluntary surrender",
      "Condition": "none",
      "CrimeStatus": "normal",
      "Confession": "truthful"
    }
  },
  "Article 347": {
    "consequences": ["fixed-term imprisonment of not more than 3 years, criminal detention, or public surveillance, and a fine"],
    "model_details": {
      "DrugQuantity": "heroin/meth < 10g",
      "Circumstance": "none",
      "Condition": "none",
      "Actor": "person",
      "Action": "selling",
      "Age": "not provided (assumed >= 18 for Article 347)"
    }
  }
}
\end{lstlisting}

\subsubsection*{Final Verdict Generated by LLM}

The final judge agent renders the verified statutes and statute-specific consequences into a natural-language judgment.

Charge: \textbf{\textit{Drug Trafficking}}\\
Judgment: \textbf{\textit{Fixed-term imprisonment of 2 years and a fine of RMB 10,000; confiscation of the illegal substances, which are to be turned over to the state treasury.}}\\
Applicable Provisions: Article 64, Article 65, Article 67, Article 347

\textbf{General Provisions:}
\begin{itemize}
\item Article 64: Illegal drugs are contraband and must be confiscated and turned over to the state treasury.
\item Article 65: Defendant committed the same crime within five years of serving a sentence, thus considered a recidivist and subject to harsher punishment.
\item Article 67: Defendant voluntarily surrendered and confessed truthfully, qualifying for leniency.
\end{itemize}

\textbf{Specific Provision:}
\begin{itemize}
\item Article 347: Selling 2g of methamphetamine constitutes low-quantity drug trafficking punishable by up to three years' imprisonment or criminal detention, plus a fine.
\end{itemize}

\textbf{Final Sentencing Outcome:} \textit{Two years' fixed-term imprisonment}

\textbf{Reasoning:} The defendant's act of selling 2g of methamphetamine meets the criteria for Article 347. As a recidivist under Article 65, he is subject to heavier punishment. However, under Article 67, his voluntary surrender and truthful confession allow for mitigation. Article 64 mandates confiscation of contraband. Thus, the final sentence includes imprisonment, fine, and confiscation.

\end{document}